\documentclass{article}

\usepackage[T1]{fontenc} %
\usepackage{iclr2027_conference,times}

\usepackage{amsmath,amsfonts,bm}

\def\1{\bm{1}}

\DeclareMathAlphabet{\mathsfit}{\encodingdefault}{\sfdefault}{m}{sl}
\SetMathAlphabet{\mathsfit}{bold}{\encodingdefault}{\sfdefault}{bx}{n}

\usepackage{hyperref}
\usepackage{url}
\usepackage{graphicx}
\usepackage{float} %
\usepackage{placeins} %
\usepackage{booktabs}
\usepackage{array}   %
\usepackage{amsmath}
\usepackage{amssymb}
\usepackage{multirow}
\usepackage{xcolor}
\usepackage{xspace}
\usepackage{etoolbox}

\newcommand{\method}{ReLift\xspace}

\title{ReLift: Dense 3D Tracking as Lifted\\Trajectory Repair}

\author{%
\makebox[\dimexpr\textwidth-2\tabcolsep\relax][c]{%
\begin{tabular}{c}
\textbf{Assaf Singer}$^{1*}$ \quad
\textbf{Maroun Khriesh}$^{1*}$ \quad
\textbf{Noam Rotstein}$^{1,2}$ \quad
\textbf{Yoni Kasten}$^{1,2}$ \quad
\textbf{Or Litany}$^{1,2}$ \\
{\normalfont $^{1}$ Technion -- Israel Institute of Technology \quad $^{2}$ NVIDIA}\\
{\normalfont $^{*}$ Equal contribution}
\end{tabular}}%
}

\iclrfinalcopy

\makeatletter
\patchcmd{\@maketitle}{\vskip 0.3in minus 0.1in}{\vskip 8pt}{}%
  {\PackageError{relift}{Could not adjust title spacing}{Check the title definition in the style file.}}
\makeatother

\begin{document}

\maketitle
\lhead{} %

\begingroup
\setlength{\intextsep}{6pt}
\renewenvironment{abstract}{%
  \vskip 14pt\centerline{\large\sc Abstract}\vspace{2pt}%
  \setlength{\topsep}{2pt}%
  \setlength{\partopsep}{0pt}%
  \begin{quote}}{\par\end{quote}\vskip 1ex}
\begin{figure}[!h]
\centering
\begin{minipage}[t]{0.52\linewidth}
\vspace{0pt}%
\centering
\includegraphics[width=\linewidth]{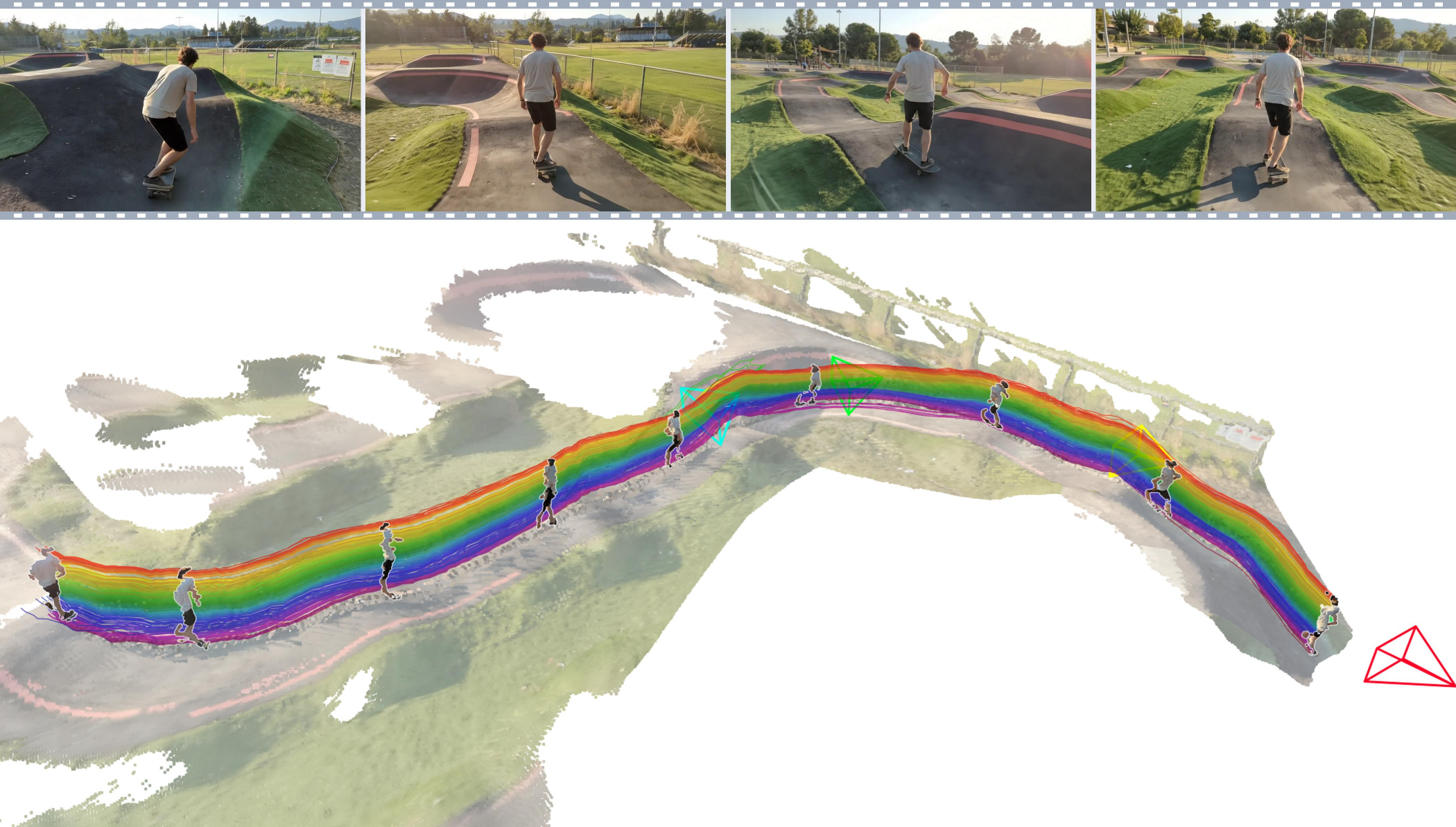}
\end{minipage}\hfill
\begin{minipage}[t]{0.47\linewidth}
\vspace{0pt}
\centering
\includegraphics[width=\linewidth]{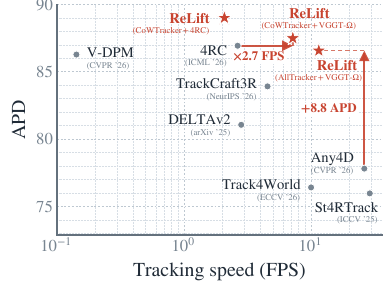}
\end{minipage}
\caption{\method combines pretrained 2D tracking and 3D geometry
with learned trajectory repair and completion, achieving state-of-the-art
dense 3D tracking accuracy on WorldTrack and a strong speed--accuracy trade-off.
\textbf{Left:} Long-range 3D trajectories recovered from a real-world video.
\textbf{Right:} WorldTrack accuracy versus full-pipeline FPS
(see Appendix~\ref{app:profiling} for details).}
\label{fig:teaser}
\end{figure}

\begin{abstract}
Dense 3D tracking from monocular video requires establishing point correspondences over time, estimating their depth, and inferring their trajectories during occlusion. 
Existing methods typically learn these tasks jointly, a demanding objective given the scarcity of 4D supervision.
We show that strong pretrained 2D trackers and 3D geometry models provide an effective foundation: lifted trajectories, with occluded segments simply interpolated, already rival dedicated 3D trackers using the same geometry. 
However, lifted trajectories inherit correspondence and depth errors, and under occlusion, their depth comes from the occluder rather than the tracked point. 
We introduce \method, an efficient feed-forward framework that recasts dense 3D tracking as geometric trajectory repair and completion. At its core, a trajectory corrector alternates attention along and across trajectories to preserve reliable observations, repair corrupted positions, and complete missing motion. Using only geometric inputs allows us to train directly on 3D trajectories with simulated lifting errors, decoupling training from images and from any specific tracker or geometry model.
Concentrating 4D supervision on trajectory refinement yields a reusable model that consistently improves lifted trajectories across trackers and geometry providers without retraining, so it benefits directly from future advances in either component. On WorldTrack, \method achieves state-of-the-art mean accuracy, improving on the strongest baselines by 2.1 APD on all points and 3.0 on dynamic points, with faster configurations remaining competitive with leading 3D trackers while running $2.7\text{--}4.4\times$ faster than 4RC.
Visit our {\hypersetup{pdfborder={0 0 0}}\href{https://relift3d.github.io/}{\textcolor{blue}{project page}}} for more examples and code.
\end{abstract}

\endgroup

\newif\ifcompactbody
\compactbodytrue
\ifcompactbody
  \setlength{\parskip}{4pt} %
  \makeatletter
  \renewcommand{\section}{\@startsection{section}{1}{\z@}%
    {-1.6ex plus -0.5ex minus -0.2ex}%
    {1ex plus 0.3ex minus 0.2ex}{\large\sc\raggedright}}
  \renewcommand{\subsection}{\@startsection{subsection}{2}{\z@}%
    {-1.4ex plus -0.5ex minus -0.2ex}%
    {0.6ex plus 0.2ex}{\normalsize\sc\raggedright}}
  \renewcommand{\paragraph}{\@startsection{paragraph}{4}{\z@}%
    {1ex plus 0.5ex minus 0.2ex}{-1em}{\normalsize\bf}}
  \makeatother
\fi

\newif\ifcompactfloats
\compactfloatstrue
\ifcompactfloats
  \setlength{\textfloatsep}{16pt plus 2pt minus 4pt} %
  \setlength{\floatsep}{10pt plus 2pt minus 2pt}     %
  \setlength{\intextsep}{10pt plus 2pt minus 2pt}    %
  \setlength{\abovecaptionskip}{7pt}                %
\fi

\section{Introduction}
\label{sec:introduction}

Dense 3D point trajectories describe how scene content moves through space over
time, and are key in downstream tasks such as manipulation policy learning,
dynamic scene reconstruction, sensor simulation, and motion-controlled video
generation~\citep{chen2025omnire,bharadhwaj2024track2act,lu2024manigaussian,geng2025motionprompting,borreda2026radargen,singer2026timetomove}.
Recovering them from monocular video is usually posed as a single learning
problem: one feed-forward network is trained end to end to output 3D tracks.

Such a network has to learn three things at once: finding correspondence across
frames, recovering the geometry of the scene, and inferring motion behind
occlusions. However, pretrained models already excel at two of these tasks.
Dedicated 2D trackers establish strong
correspondences~\citep{harley2025alltracker,lai2026cowtracker} and can exceed
the image-plane accuracy of projected 3D predictions (Appendix~\ref{app:apd2d}),
while geometry models provide depth and cameras learned from scalable geometric
supervision without requiring annotated dynamic 4D
trajectories~\citep{lin2025depth,wang2026vggtomega}.
This motivates focusing the scarce 4D supervision on what remains: repairing
errors in the combined estimates and completing occluded motion.

We therefore introduce \method, which combines strong 2D correspondences and
pretrained scene geometry for dense 3D tracking.
We take the viewing ray of each point from a 2D tracker and its
range along that ray from a depth model, and lift the two into a 3D trajectory
we call the lift.
Since occluded tracks sample the occluder's depth, we discard predictions flagged as occluded and initialize missing positions by interpolation.
Even before learned correction, these initialized lifts can match or exceed the mean accuracy of
learned 3D trackers using the same geometry (Tables~\ref{tab:main_apd} and~\ref{tab:ablation}).
However, these lifts still lack observations during occlusion and suffer from jumps at depth
boundaries and jitter from per-frame depth fluctuations. These failures can produce severe
trajectory discontinuities and leave hidden motion unresolved.
We recast dense 3D tracking as geometric trajectory repair and completion.
A compact learned corrector preserves reliable observations while repairing corrupted positions and
recovering missing motion. Alternating attention along and across tracks shares temporal and spatial
evidence, while processing track groups independently keeps dense feed-forward correction efficient.

A natural training strategy is to run pretrained tracking and geometry models on synthetic videos
with ground-truth 3D trajectories, then train the corrector to recover those trajectories from the
resulting lifts. However, this ties the training inputs to the errors of the chosen estimators.
We avoid this by training without ever running a tracker, a depth model, or a
renderer. We model the geometric effects of lifting failures directly as
randomized corruptions of clean 3D trajectories from
PointOdyssey~\citep{zheng2023pointodyssey}, producing \emph{simulated lifts} from which
we train a network to recover the clean trajectories.
Training uses geometric annotations without images or provider predictions,
concentrating supervision on learning motion priors for trajectory repair and
completion (Sec.~\ref{sec:training}).
Our ablation supports this choice: training on simulated lifts improves accuracy over
training on lifts generated by AllTracker and Depth Anything~3 (Sec.~\ref{sec:exp_ablation}).

Training on simulated lifts supports transfer across providers. We evaluate the same \method
corrector without retraining with AllTracker~\citep{harley2025alltracker} or
CoWTracker~\citep{lai2026cowtracker}, each paired with depth and camera estimates from
Depth Anything~3~\citep{lin2025depth}, VGGT-$\Omega$~\citep{wang2026vggtomega}, or 4RC~\citep{luo20264rc}.
Correction improves all six lifts (Fig.~\ref{fig:gain}).
Moreover, using the same depth and camera estimates as 4RC, our 2D tracking and correction pipeline
achieves higher mean tracking accuracy than 4RC's trajectory predictions (Table~\ref{tab:main_apd}).

Our contributions can be summarized as follows:
\begin{itemize}
\item \textbf{A strong lifting baseline.} We show that pretrained 2D tracks lifted with
estimated geometry and simple gap initialization can match or exceed dedicated 3D trackers
using the same geometry.
\item \textbf{\method,} an efficient feed-forward framework for dense monocular 3D tracking that
combines pretrained 2D motion and 3D geometry through learned trajectory repair and completion.
\item A provider-agnostic training strategy based on randomized geometric corruptions of synthetic trajectories,
without running tracking or geometry models during training. The same weights transfer
across evaluated providers without retraining, allowing the framework to incorporate future advances in
either component.
\item Our framework achieves state-of-the-art mean dense 3D tracking accuracy on WorldTrack, with faster
configurations retaining competitive accuracy at substantially lower runtime than comparably
accurate 3D trackers.%
\end{itemize}

\section{Related Work}
\label{sec:related_work}

\paragraph{Dynamic 3D reconstruction from monocular video.}
Classical non-rigid structure-from-motion methods recover time-varying 3D
structure from image correspondences using local rigidity and low-rank
constraints to reduce monocular ambiguity~\citep{bregler2000recovering,akhter2008nonrigid,taylor2010nonrigid,dai2014simple}.
Modern per-video optimization methods combine image evidence with depth,
tracking, and deformation priors~\citep{zhang2022structure,liu2023robust}.
Shape of Motion~\citep{wang2025shapeofmotion} and MoSca~\citep{lei2024mosca}
combine pretrained depth and tracking cues with structured motion
representations for reconstruction and rendering.
These cues provide the initialization that rendering-based optimization needs, since
photometric gradients vanish when rendered and observed content do not
overlap~\citep{cohen2026spectralsplats}.
Uni4D~\citep{yao2025uni4d} and TrackingWorld~\citep{lu2025trackingworld} optimize camera poses and
dense 3D trajectories from pretrained depth, tracking and segmentation models, or from densified
2D tracks. These methods demonstrate the strength
of combining 2D motion with 3D geometry, but require lengthy per-scene optimization;
we pursue this combination with a feed-forward model.

\paragraph{Feed-forward 3D geometry models.}
DUSt3R and MASt3R predict geometry and correspondences from image
pairs~\citep{wang2024dust3r,leroy2024mast3r}, while VGGT, VGGT-$\Omega$, and
Depth Anything~3 extend geometry estimation to multiple
views~\citep{wang2025vggt,wang2026vggtomega,lin2025depth}.
MonST3R adapts point-map prediction to dynamic scenes and aligns predictions across a
video~\citep{zhang2025monst3r}; St4RTrack and Dynamic
Point Maps also represent motion and correspondences~\citep{feng2025st4rtrack,sucar2025dynamic},
with V-DPM extending pairwise dynamic point maps to video~\citep{sucar2026vdpm}.
Geometry and tracking capabilities thus overlap. In our pipeline, frozen geometry models
supply depth and cameras, while 2D trackers establish point identities across frames.

\paragraph{2D tracking.}
2D tracking follows the image projections of scene points over time.
CoTracker jointly tracks multiple points~\citep{karaev2024cotracker}, while AllTracker and
CoWTracker extend tracking to every reference-frame pixel~\citep{harley2025alltracker,lai2026cowtracker}.
Per-video self-supervision also yields strong correspondences without ground-truth annotations
for the input video:
DINO-Tracker adapts pretrained image features~\citep{tumanyan2024dinotracker}, while OmniMotion
optimizes globally consistent tracks~\citep{wang2023omnimotion}.
These approaches show that correspondence estimation can exploit real-video evidence without
annotated 3D motion, motivating its use as a separate component.
Our comparison supports reusing
strong 2D correspondences: native CoWTracker surpasses the evaluated 3D methods in image-space
accuracy (Appendix~\ref{app:apd2d}).

Established tracks also provide a useful representation for recovering dynamic 3D structure.
TracksTo4D reconstructs geometry and motion from 2D tracks using a feed-forward, low-rank
formulation without 3D supervision~\citep{kasten2024tracksto4d}. Seurat learns depth changes from
2D trajectories and combines them with pretrained depth estimates~\citep{cho2025seurat}.
Building on these precedents, \method starts from tracks already lifted with pretrained depth and
cameras, and learns 3D residuals to repair corrupted observations and complete missing motion.

\paragraph{Dense feed-forward 3D tracking.}
Feed-forward 3D trackers estimate trajectories from video, either jointly predicting geometry and
motion or conditioning on externally supplied depth.
DELTA and DELTAv2 predict dense long-range 3D tracks using video and depth
inputs~\citep{ngo2025delta,ngo2025deltav2}, while SpatialTrackerV2 jointly
estimates geometry, camera motion, and point motion~\citep{xiao2025spatialtracker}.
Trace Anything, Track4World, and 4RC predict dense trajectories or queryable
geometry and motion~\citep{liu2025traceanything,lu2026track4world,luo20264rc}.
TrackCraft3R repurposes pretrained video-diffusion models for dense 3D
tracking~\citep{nam2026trackcraft3r}; MotionCrafter uses video generative
priors for geometry and scene flow~\citep{zhu2026motioncrafter}.
Available 3D motion supervision includes complete synthetic trajectories from
PointOdyssey~\citep{zheng2023pointodyssey} and trajectories reconstructed from real stereo videos
in Stereo4D~\citep{jin2025stereo4d}. We use PointOdyssey to learn repair and completion of
lifted trajectories. Under matched geometry, these lifts already rival dedicated 3D trackers,
and correction further improves them
(Sec.~\ref{sec:exp_main}).

\section{Method}
\label{sec:method}

Given a monocular video, we recover dense 3D trajectories in the camera coordinates of the reference frame $I_0$.
Frozen pretrained models supply 2D tracks, depth and camera estimates, which we combine into
lifted trajectories (Sec.~\ref{sec:lift}).
The lift is accurate where tracking and depth are reliable.
We use predicted visibility to mask observations, leaving gaps to complete. Retained observations
may still be corrupted by missed occlusions, depth errors or tracking drift, and therefore require repair.

\begin{figure}[t]
\centering
\includegraphics[width=0.99\linewidth]{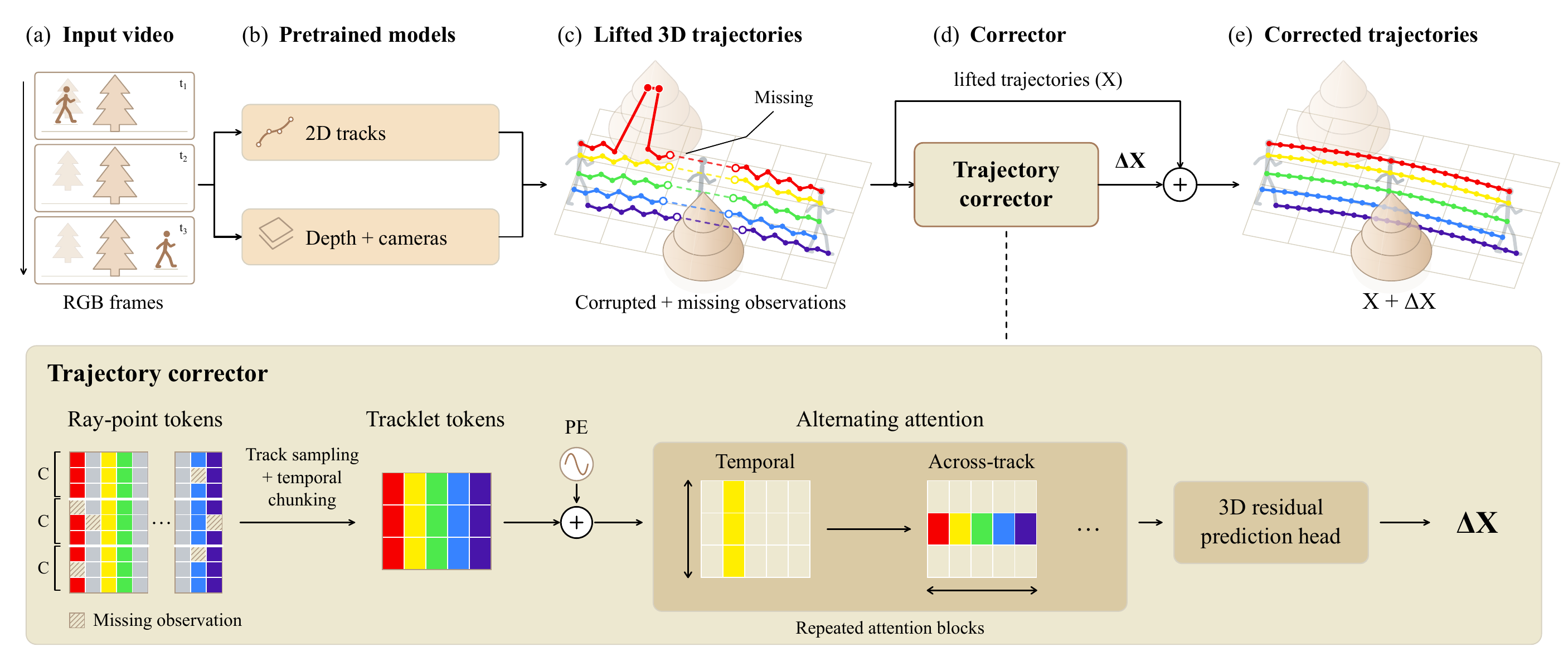}
\caption{\textbf{Inference pipeline of \method.} Frozen pretrained models lift video observations
(a--b) into corrupted and incomplete 3D trajectories (c). The corrector adds residuals to
initialized XYZ coordinates to produce the final trajectories (d--e). Below: ray--point tokens
$\boldsymbol\phi_{ti}$ are grouped into $C$-frame tracklets and embedded as tokens $\mathbf h_{ni}$,
then processed by alternating temporal and across-track attention. The output head predicts
frame-level XYZ residuals, which are added to the initialized trajectories. We train the corrector
on simulated lifts from clean trajectory annotations (Sec.~\ref{sec:training}).}
\label{fig:overview}
\end{figure}

Our corrector model uses positions, viewing rays and availability masks (Sec.~\ref{sec:representation}).
We hypothesize that temporal inconsistencies and disagreements between tracks provide cues for repair, while observations before and after gaps, together with related motion, can guide completion.
We therefore use alternating attention along and across tracks to aggregate these cues and predict 3D residual corrections (Sec.~\ref{sec:corrector}).
Temporal chunking and batched track groups enable efficient
feed-forward correction over all reference pixels (Fig.~\ref{fig:overview}).

Because the corrector uses only geometry, we train it on simulated lifts
constructed directly from clean 3D trajectory annotations, without rendering
images or running providers (Sec.~\ref{sec:training}).

\subsection{Lifting Tracks into 3D}
\label{sec:lift}

For a video $\{I_t\}_{t=0}^{T-1}$, a dense 2D tracker provides image trajectories
$\mathbf u_{ti}$ and visibility scores $v_{ti}\in[0,1]$ for pixels $i$ of the reference frame $I_0$.
A geometry model provides optical-axis depth maps $Z_t$, intrinsics $K_t$ and camera poses.
We express geometry in the reference camera coordinate system: $R_t\in\mathrm{SO}(3)$ rotates from camera $t$ to
camera $0$, and $\mathbf o_t$ is camera $t$'s center in that coordinate system, with $R_0=I$ and
$\mathbf o_0=\mathbf 0$.
For homogeneous
image coordinates $\widetilde{\mathbf u}_{ti}=(\mathbf u_{ti},1)$, let
$\mathbf q_{ti}=K_t^{-1}\widetilde{\mathbf u}_{ti}$. The unit viewing direction, range and lifted
position are
\begin{equation}
  \mathbf d_{ti}=\frac{R_t\,\mathbf q_{ti}}{\|\mathbf q_{ti}\|},\qquad
  r_{ti}=Z_t(\mathbf u_{ti})\,\|\mathbf q_{ti}\|,\qquad
  \mathbf y_{ti}=\mathbf o_t+r_{ti}\,\mathbf d_{ti}.
  \label{eq:lift}
\end{equation}
Since $\mathbf q_{ti}$ has third coordinate $1$, its norm converts optical-axis depth into distance along the ray.

We distinguish four common failure patterns in lifted trajectories.
\emph{Correspondence errors} introduced by the 2D tracker perturb the viewing direction.
\emph{Depth jitter and drift} displace points along the viewing ray.
Jitter produces rapid frame-to-frame fluctuations in depth error. Drift is a gradual
change in the difference between estimated and true depth over successive frames.
\emph{Wrong-surface observations} arise when the sampled depth belongs to another
surface, for example near a depth boundary or when a track follows an occluder.
\emph{Missing observations} can arise when a point is occluded or the 2D tracker
loses track of it.

We use the tracker's predicted visibility to determine observation availability.
We retain a lifted position and its ray direction when its range is positive and the tracker's
visibility score exceeds a tracker-specific threshold $\gamma$; otherwise, we
mask both. The binary mask $m_{ti}$ records this choice, with $m_{0i}=1$ for
valid reference points. Correctly detected occlusions therefore produce gaps,
while missed occlusions can leave observations on the occluder's surface.
Observations before and after a gap retain the same track index.
Preprocessing details are given in Appendix~\ref{app:observation_preprocessing},
and Appendix~\ref{app:ray_masking} evaluates the choice to mask ray directions.

\subsection{Trajectory Representation}
\label{sec:representation}

\paragraph{Normalization and initialization.}
We normalize positions and camera centers by the median reference depth $s$ and initialize
each trajectory with the normalized lift, $\bar{\mathbf x}_{ti}=\mathbf y_{ti}/s$, at available
observations.
Missing segments are filled by linear interpolation between observations; after
the last observation, we hold its position.

\paragraph{Ray--point tokenization.}
For each track and frame, we construct a token combining initialized displacement,
retained observations and viewing geometry. Let $\mathbf a_i=\mathbf y_{0i}/s$ be
the track's normalized initial 3D position (the reference anchor), and $g_{ti}$
the distance in frames to the nearest available observation. The token is
\begin{equation}
  \boldsymbol\phi_{ti}=\bigl(\bar{\mathbf x}_{ti}-\mathbf a_i,\;
  m_{ti}\,(\bar{\mathbf x}_{ti}-\mathbf a_i),\; m_{ti},\; g_{ti},\;
  m_{ti}\,\mathbf d_{ti},\; \mathbf o_t/s\bigr)\in\mathbb R^{14}.
  \label{eq:input}
\end{equation}
The first displacement provides an estimate at every frame, including gaps.
The second exposes only retained observations: it equals
$\bar{\mathbf x}_{ti}-\mathbf a_i$ when $m_{ti}=1$ and is zero otherwise.
The mask indicates whether an observation is available, while $g_{ti}$ measures
the temporal distance to the nearest available observation.
The ray identifies the direction of range errors and is masked inside gaps;
the camera center supplies its origin.

For efficiency, we concatenate non-overlapping chunks of $C$ consecutive
ray--point tokens from each track and project each chunk into a single
\emph{tracklet token} for attention. This gives $N=\lceil T/C\rceil$ tokens
per track, padding the final chunk when necessary. For the frame indices
$\mathcal T_n$ of chunk $n$,
\begin{equation}
  \mathbf h_{ni}=W_{\mathrm{in}}
  \bigl[\boldsymbol\phi_{ti}\bigr]_{t\in\mathcal T_n}
  +\mathbf p_n+\boldsymbol\psi(\mathbf a_i)\in\mathbb R^D.
  \label{eq:token}
\end{equation}
Here $W_{\mathrm{in}}\in\mathbb R^{D\times14C}$ projects the concatenated
features to width $D$, and $\mathbf p_n$ encodes the chunk's temporal position.
The anchor embedding $\boldsymbol\psi(\mathbf a_i)$ encodes the track's initial
3D location, providing spatial context alongside the anchor-relative
displacements. Additional details are in
Appendix~\ref{app:corrector_implementation}.

\subsection{Trajectory Corrector}
\label{sec:corrector}

To limit attention cost, the corrector processes groups of $P$ tracks spanning all $T$ frames.
Their tracklet tokens form a grid $H\in\mathbb R^{P\times N\times D}$, with one row per track
and one column per chunk.

We process $H$ with alternating temporal and across-track self-attention, following the
factorization of TracksTo4D~\citep{kasten2024tracksto4d}. Temporal attention connects all $N$ chunks
of each track, aggregating evidence across the full sequence along the correspondences supplied
by the 2D tracker. Across-track attention connects the $P$ tracks within each temporal chunk,
sharing motion information across the group. For fixed feature width,
attention costs $O(PN^2+NP^2)$ per block, compared with $O(P^2N^2)$ for joint attention over all tokens.

A prediction head maps each tracklet token to $C$ XYZ residuals $\boldsymbol\Delta_{ti}$,
one per frame. Viewing rays condition these predictions, but the residuals are unrestricted in 3D,
allowing correction of range and correspondence errors. We add them to the initialized
trajectories and restore the original scale:
\begin{equation}
  \hat{\mathbf x}_{ti}=s\,\bigl(\bar{\mathbf x}_{ti}+\boldsymbol\Delta_{ti}\bigr).
  \label{eq:output}
\end{equation}
We retain the tracker's visibility scores $v_{ti}$ unchanged.

\paragraph{Efficient dense inference.}
\label{sec:inference}
At inference, we randomly partition the reference pixels into groups of $P$ tracks and process the
groups independently in batches. All groups use the clip's shared scale $s$, and their predictions
are assembled into the dense trajectory field at the corresponding reference pixels. For fixed
sequence length and group size, the corrector's cost grows linearly with the number of pixels.
Implementation details and scaling measurements are given in Appendices~\ref{app:dense_inference}
and~\ref{app:profiling}.

\subsection{Training from Simulated Lifts}
\label{sec:training}

We train the corrector on  3D trajectories from the PointOdyssey training
split~\citep{zheng2023pointodyssey}.
Each sample is a group of trajectories
$\mathbf x_{ti}$ over $T$ frames, ground-truth visibility $v^\star_{ti}\in\{0,1\}$, and camera centers
$\mathbf o_t$ in reference camera coordinates. We generate simulated lifts directly from these
annotations, without images or provider inference, so lifting errors can be varied independently
of any tracker or geometry model.

\paragraph{Simulated lifting errors.}
We simulate lifting failures caused by imperfect 2D trackers and 3D geometry
models (Sec.~\ref{sec:lift}) by perturbing viewing ray directions, ranges and
observation availability. For each clean point
$\mathbf x_{ti}$, its range
$r^\star_{ti}=\|\mathbf x_{ti}-\mathbf o_t\|$ is its distance from the camera,
and $\mathbf d^\star_{ti}=(\mathbf x_{ti}-\mathbf o_t)/r^\star_{ti}$ is its
unit viewing direction. We construct
\begin{equation}
  \mathbf d_{ti}=Q_{ti}\mathbf d^\star_{ti},\qquad
  r_{ti}=r^\star_{ti}\exp(\varepsilon_{ti}),\qquad
  \mathbf y_{ti}=\mathbf o_t+r_{ti}\,\mathbf d_{ti}.
  \label{eq:simulated_lift}
\end{equation}
The rotation $Q_{ti}\in\mathrm{SO}(3)$ applies small angular perturbations to
approximate \emph{correspondence errors} introduced by the 2D tracker. The log-range
perturbation $\varepsilon_{ti}$ uses independent noise to model \emph{depth jitter}
and temporally correlated offsets to approximate \emph{depth drift}.
Occasional large offsets additionally approximate \emph{wrong-surface observations}.
The exponential converts these log-range offsets to multiplicative range changes.
Camera centers remain unchanged; we do not explicitly simulate camera-pose
errors or persistent switches to another object's trajectory.

To generate \emph{missing observations}, we initialize the input mask $m_{ti}$
from ground-truth visibility $v^\star_{ti}$ and remove observations using moving
geometric occluders and contiguous temporal dropouts. To model wrong-surface
observations caused by missed occlusions, we retain some observations near the
start or end of a synthetic occlusion ($m_{ti}=1$) at ranges corresponding to
the occluder's depth.
Clean trajectories remain the targets at every frame, including those with
$m_{ti}=0$. The simulated observations undergo the same normalization,
initialization and tokenization as at inference
(Sec.~\ref{sec:representation}), with scale and anchors computed from the
corrupted input. Appendix~\ref{app:training_corruptions} specifies the
corruption distributions, and Sec.~\ref{sec:exp_ablation} evaluates their
contributions.

\paragraph{Training for transfer.}
To encourage transfer across trackers and geometry models, we vary the
parameters of these corruption processes across training examples, following
domain randomization in
robotics~\citep{tobin2017domainrandomization,peng2018dynamicsrandomization}.
We sample quantities such as noise strengths, temporal correlation
coefficients and outlier rates from fixed distributions
(Appendix~\ref{app:training_corruptions}).
This exposes the corrector to diverse patterns of lifting errors without
using provider-generated training inputs. The same trained corrector
improves the initialized lifts of all six evaluated tracker--geometry
combinations (Fig.~\ref{fig:gain}).

\paragraph{Reconstruction objective.}
We supervise 3D positions and frame-to-frame displacements, including where input observations are
unavailable, with a weighted squared-error objective,
\begin{equation}
  \mathcal L=\frac{\sum_{t,i}w_{ti}\|\mathbf e_{ti}\|^2}{\sum_{t,i}w_{ti}}
  +\lambda\,\frac{\sum_{t\ge1,i}w_{ti}\|\mathbf e_{ti}-\mathbf e_{t-1,i}\|^2}{\sum_{t\ge1,i}w_{ti}}.
  \label{eq:loss}
\end{equation}
Here $\mathbf e_{ti}=(\hat{\mathbf x}_{ti}-\mathbf x_{ti})/(s\rho_i)$ is the relative position
error, with $\rho_i=\max(\|\mathbf a_i\|,\rho_{\min})$ the floored reference range in normalized
coordinates. This scaling prevents distant points from dominating the objective. The first term
supervises positions; the second, weighted by $\lambda$, matches predicted and true displacements
between consecutive sampled frames, discouraging jitter while preserving the target motion.
The weights $w_{ti}$ count moving tracks four times as much as static ones, and unavailable observations
twice as much as available ones (Appendix~\ref{app:training}).

\section{Experiments}
\label{sec:experiments}

\subsection{Experimental Setup}
\label{sec:exp_setup}

\paragraph{Evaluation Datasets.}
We evaluate dense 3D point tracking in world coordinates on the four
WorldTrack~\citep{feng2025st4rtrack} datasets: the real-world Aria Digital Twin (ADT)~\citep{pan2023adt}
and Panoptic Studio (PS)~\citep{joo2019panopticstudio} from TAPVid-3D~\citep{koppula2024tapvid3d}, and
the synthetic PointOdyssey (PO)~\citep{zheng2023pointodyssey} and Dynamic
Replica (DR)~\citep{karaev2023dynamicstereo}. We score dense predictions at the sparse ground-truth
trajectories of points visible in the reference frame, over the first 84 frames.
We also evaluate on TrackCraft3R's~\citep{nam2026trackcraft3r} 50 24-frame Kubric~\citep{greff2022kubric}
clips, with ground-truth trajectories for every reference pixel
(Appendix~\ref{app:kubric}).

\paragraph{Metrics.}
We report the Average Percentage of 3D Points within Delta (APD3D$\uparrow$)~\citep{koppula2024tapvid3d},
averaged over distance thresholds $\delta\in\{0.1,0.3,0.5,1.0\}$\,m. Following
4RC~\citep{luo20264rc}, we evaluate all points (\emph{All}) and dynamic points (\emph{Dyn}) after
RANSAC Sim(3) alignment, fitted separately to each scoring set. We also evaluate image-plane
accuracy using APD2D (Appendix~\ref{app:apd2d}). Appendices~\ref{app:evaluation} and~\ref{app:experiments}
provide scoring details, complementary metrics (AJ, OA and EPE), and global-scale alignment results.

\paragraph{Baselines.}
We pair AllTracker~\citep{harley2025alltracker} or
CoWTracker~\citep{lai2026cowtracker} with depth and camera estimates from
Depth Anything~3 (DA3)~\citep{lin2025depth} or
VGGT-$\Omega$~\citep{wang2026vggtomega}.
TrackCraft3R~\citep{nam2026trackcraft3r} and
DELTAv2~\citep{ngo2025deltav2} receive the same estimated depth,
cameras and intrinsics as ours.
We also pair both 2D trackers with 4RC's~\citep{luo20264rc} own depth
and camera estimates, comparing our tracking and correction pipeline
with 4RC's trajectory predictions under matched geometry.
Additional baselines include V-DPM~\citep{sucar2026vdpm},
St4RTrack~\citep{feng2025st4rtrack}, Any4D~\citep{karhade2025any4d},
Track4World~\citep{lu2026track4world},
TraceAnything~\citep{liu2025traceanything},
MotionCrafter~\citep{zhu2026motioncrafter}, and
OpenD4RT~\citep{rhos2026opend4rt}, an open reimplementation of
D4RT~\citep{zhang2025d4rt}.
No method receives ground-truth intrinsics
(Appendices~\ref{app:method} and~\ref{app:baselines}).

\subsection{Quantitative Results}
\label{sec:exp_main}

\begin{table}[t]
\centering
\caption{\textbf{Dense 3D tracking on WorldTrack.} APD3D (\%) on all (All) and dynamic (Dyn) points
after RANSAC Sim(3) alignment; bold best, underline second best. Time includes the complete pipeline.
\textsuperscript{\dag}Timing at $256^2$ instead of $512^2$.
\textsuperscript{\ddag}Restricted ADT coverage for MotionCrafter, also affecting its means
(Appendix~\ref{app:baselines}).}
\label{tab:main_apd}
\smallskip
\resizebox{\linewidth}{!}{%
\begin{tabular}{@{}lccccccccccc@{}}
\toprule
 & \multicolumn{2}{c}{ADT} & \multicolumn{2}{c}{PS} & \multicolumn{2}{c}{PO} & \multicolumn{2}{c}{DR} & \multicolumn{2}{c}{Mean} &  \\
\cmidrule(lr){2-3}\cmidrule(lr){4-5}\cmidrule(lr){6-7}\cmidrule(lr){8-9}\cmidrule(lr){10-11}
Method & All & Dyn & All & Dyn & All & Dyn & All & Dyn & All & Dyn & Time (s) \\
\midrule
4RC & 87.07 & 78.57 & 86.25 & 86.10 & \textbf{85.92} & 81.17 & 88.51 & 80.56 & 86.94 & 81.60 & 31.8 \\
Any4D & 76.17 & 78.32 & 77.57 & 77.61 & 73.96 & 73.56 & 83.54 & 72.83 & 77.81 & 75.58 & 3.2 \\
MotionCrafter & 67.04\textsuperscript{\ddag} & 83.34\textsuperscript{\ddag} & 62.60 & 62.63 & 46.29 & 49.09 & 70.83 & 62.39 & 61.69\textsuperscript{\ddag} & 64.36\textsuperscript{\ddag} & 20.4 \\
OpenD4RT & 71.71 & 79.68 & 84.36 & 85.27 & 68.39 & 74.21 & 77.55 & 79.69 & 75.50 & 79.71 & 284\textsuperscript{\dag} \\
St4RTrack & 81.39 & 79.39 & 72.84 & 72.24 & 71.36 & 71.44 & 78.29 & 67.23 & 75.97 & 72.58 & 2.9 \\
Track4World & 77.48 & 91.80 & 80.10 & 79.39 & 69.13 & 69.43 & 78.94 & 71.98 & 76.41 & 78.15 & 8.4 \\
TraceAnything & 81.76 & 77.01 & 71.88 & 71.66 & 49.77 & 53.14 & 68.91 & 61.80 & 68.08 & 65.90 & 5.2 \\
V-DPM & 88.42 & 79.40 & 89.26 & 89.49 & 84.46 & \textbf{85.95} & 82.99 & 80.80 & 86.28 & 83.91 & 591 \\
\midrule
DELTAv2 + DA3 & 83.63 & 92.19 & 85.38 & 85.37 & 66.32 & 67.35 & 76.88 & 64.50 & 78.06 & 77.35 & 33.4 \\
DELTAv2 + VGGT-$\Omega$ & 84.57 & 91.67 & 89.41 & 89.22 & 67.49 & 68.60 & 82.79 & 72.23 & 81.06 & 80.43 & 29.8 \\
TrackCraft3R + DA3 & 91.75 & 90.82 & 84.39 & 83.95 & 74.04 & 71.42 & 78.44 & 65.84 & 82.16 & 78.01 & 22.1 \\
TrackCraft3R + VGGT-$\Omega$ & 92.56 & 89.54 & 86.56 & 86.65 & 72.66 & 71.48 & 83.89 & 73.74 & 83.92 & 80.36 & 18.5 \\
\midrule
\multicolumn{12}{@{}l}{\emph{Ours, one trained corrector for all rows} (2D tracker + geometry provider)} \\
\textbf{AllTracker + DA3} & 93.69 & \textbf{94.98} & 85.46 & 85.59 & 76.04 & 77.44 & 79.50 & 68.94 & 83.68 & 81.74 & 11.4 \\
\textbf{AllTracker + VGGT-$\Omega$} & \underline{94.04} & 92.17 & \underline{90.40} & \underline{90.46} & 76.10 & 78.96 & 85.74 & 78.10 & 86.57 & 84.92 & 7.3 \\
\textbf{AllTracker + 4RC} & 91.07 & 94.13 & 88.73 & 88.65 & 83.63 & 80.26 & \underline{88.81} & \underline{81.08} & \underline{88.06} & \underline{86.03} & 36.4 \\
\textbf{CoWTracker + DA3} & 93.78 & \underline{94.85} & 86.99 & 86.64 & 77.63 & 79.19 & 79.72 & 68.19 & 84.53 & 82.22 & 15.3 \\
\textbf{CoWTracker + VGGT-$\Omega$} & \textbf{94.14} & 92.09 & \textbf{92.20} & \textbf{92.06} & 77.63 & 80.49 & 86.00 & 78.70 & 87.49 & 85.84 & 11.7 \\
\textbf{CoWTracker + 4RC} & 91.41 & 94.31 & 89.97 & 89.83 & \underline{85.60} & \underline{82.14} & \textbf{89.05} & \textbf{81.49} & \textbf{89.01} & \textbf{86.95} & 40.4 \\
\bottomrule
\end{tabular}}

\end{table}

\paragraph{Accuracy.}
Our method achieves state-of-the-art mean APD3D on all and dynamic points with CoWTracker and
4RC geometry (Table~\ref{tab:main_apd}). It is particularly strong on the real-world datasets ADT and PS, where our variants lead both metrics.
All six configurations share one trained corrector: for both trackers, mean accuracy
increases from DA3 to VGGT-$\Omega$ to 4RC geometry, demonstrating gains from swapping geometry
providers without retraining.

\paragraph{Combining 2D motion and 3D geometry.}
To test the decomposition introduced in Sec.~\ref{sec:introduction}, we compare methods with
the same geometry inputs. Even before correction, CoWTracker lifts with initialized gaps and
VGGT-$\Omega$ geometry reach 84.9 mean all-point APD3D (Table~\ref{tab:ablation}, first row),
compared with 83.9 for TrackCraft3R and 81.1 for DELTAv2 using the same geometry
(Table~\ref{tab:main_apd}).
After correction, our method outperforms both in every APD3D column of Table~\ref{tab:main_apd}.
Given 4RC's own geometry, our CoWTracker variant improves on 4RC's mean accuracy by 2.1 (All) and
5.4 (Dyn) points, demonstrating the advantage of our pipeline with geometry held fixed.
On Kubric, our DA3 and VGGT-$\Omega$ variants trail 4RC but lead methods sharing their geometry
on all-point APD3D; with 4RC geometry, ours outperform 4RC on dynamic points
(Appendix~\ref{app:kubric}).

\begin{figure}[t]
\centering
\includegraphics[width=\linewidth]{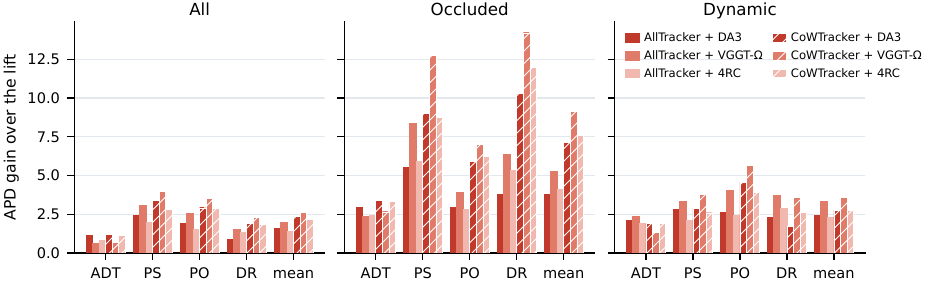}
\caption{\textbf{Gains from trajectory repair and completion.}
APD3D gains (\%) over the initialized lift
(Sec.~\ref{sec:representation}), per dataset and averaged, for all six
tracker--geometry combinations; hatching denotes CoWTracker.
\emph{Occluded} selects ground-truth-occluded point--frames from
\emph{All}, retaining the All-point alignment. This subset may differ
from the corrector's masked inputs.}
\label{fig:gain}
\end{figure}

\paragraph{Gains from trajectory repair and completion.}
Figure~\ref{fig:gain} measures what correction adds to the initialized lift
(Sec.~\ref{sec:representation}). Accuracy improves across datasets and all six provider combinations,
with the largest gains on ground-truth-occluded points. Correction adds 1.6--3.6 points to mean APD3D
in Table~\ref{tab:ablation}. With CoWTracker and VGGT-$\Omega$, all-point APD3D rises from
84.9 to 87.5, and the corrected model surpasses 4RC and V-DPM on all and dynamic points.
\paragraph{Image-plane accuracy.}
\label{sec:exp_apd2d}
Native CoWTracker outperforms the projected trajectories of every evaluated 3D method in APD2D,
on each dataset for both all and dynamic points (Table~\ref{tab:main_apd2d}), supporting our use
of strong pretrained 2D correspondences. Among 3D methods, our CoWTracker with 4RC geometry
leads mean all-point APD2D and ranks second on dynamic points, alongside the best mean APD3D
on both sets. Our projections remain below native CoWTracker; Appendix~\ref{app:apd2d} analyzes
this gap and the distinction between image-plane and depth accuracy.

\paragraph{Efficiency.}
Our VGGT-$\Omega$ configurations offer competitive accuracy while running $2.7$--$4.4\times$
faster than 4RC, including all provider inference (Table~\ref{tab:main_apd}). Using 4RC geometry
requires more time but delivers our highest mean accuracy. Appendix~\ref{app:profiling} analyzes
runtime and memory across resolutions.

\subsection{Qualitative Results}
\label{sec:exp_qualitative}

\begin{figure}[t]
\centering
\includegraphics[width=\linewidth]{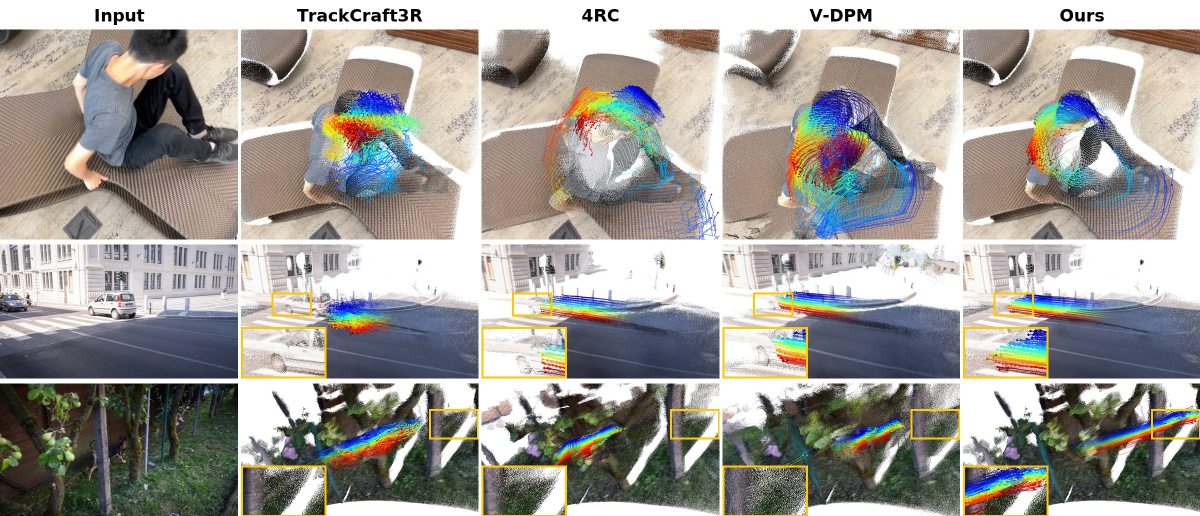}
\caption{\textbf{Qualitative comparison.} Top to bottom:
DyCheck \emph{spin}~\citep{gao2022monocular} and DAVIS \emph{car-shadow}
and \emph{libby}~\citep{perazzi2016benchmark}.
Trajectories of the same reference pixels are shown from matched novel viewpoints.
Ours uses AllTracker and VGGT-$\Omega$; TrackCraft3R shares this geometry,
while 4RC and V-DPM use their own reconstructions. Yellow boxes mark
magnified regions.}
\label{fig:qualitative}
\end{figure}

Figure~\ref{fig:qualitative} illustrates tracking under rotational motion,
large displacements and occlusion. In \emph{spin}, our trajectories form
coherent arcs, while TrackCraft3R produces scattered paths.
Our tracks follow the car through shadow-induced appearance changes in
\emph{car-shadow} and recover the dog's head after occlusion in
\emph{libby}, where competing predictions stop short.
Additional qualitative comparisons in Appendix~\ref{app:qualitative} include the same initialized
lift before and after correction, showing reduced scatter and more coherent motion with tracks
and geometry held fixed. These comparisons complement the quantitative gains in Fig.~\ref{fig:gain}.
Videos are available on our {\hypersetup{pdfborder={0 0 0}}\href{https://relift3d.github.io/}{\textcolor{blue}{project page}}}.

\subsection{Ablation Study}
\label{sec:exp_ablation}

\begin{figure}[t]
\centering
\includegraphics[width=0.84\linewidth]{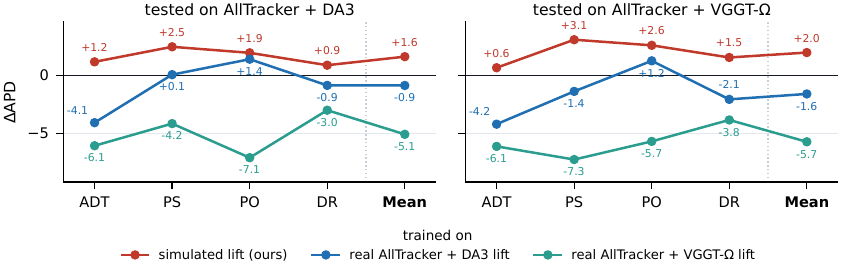}
\caption{\textbf{Simulated versus provider-generated lifts.}
APD3D gains on all points, in percentage points over the initialized lift,
with AllTracker and DA3 (left) or VGGT-$\Omega$ (right).
``Real'' denotes non-simulated, provider-generated training lifts.
All runs use the same architecture and 150{,}000 training steps.
Appendix~\ref{app:ablation_controls} reports a control with matched training windows.}
\label{fig:regime_datasets}
\end{figure}

\paragraph{Simulated versus provider-generated lifts.}
We test whether training on provider-generated lifting errors transfers
better than training on simulated corruptions. Both approaches use
PointOdyssey with the same network and training budget
(Fig.~\ref{fig:regime_datasets}).
A single simulated-lift model improves over the initialized lift on every
dataset with both test geometry providers, outperforming models trained on
AllTracker+DA3 or AllTracker+VGGT-$\Omega$ lifts.
Both provider-trained models improve on their training scenes
($+14.1$ and $+8.3$ APD3D points for DA3 and VGGT-$\Omega$, respectively),
but these gains transfer inconsistently. DA3-lift training improves accuracy on
held-out PointOdyssey scenes, but simulated training yields larger gains.
The model trained on VGGT-$\Omega$ lifts reduces accuracy on every
evaluated dataset.

Provider-generated training requires aligning ground-truth targets, and
discarding windows with unreliable alignment (Appendix~\ref{app:ablation_controls}).
 To control for
this selection, we repeat simulated-lift training on the same windows
retained for DA3-lift training. Its advantage persists with both test geometry providers
(Fig.~\ref{fig:regime_matched_data}), showing that differences in the
selected windows alone do not explain the result.
Simulation also avoids provider inference and target alignment during
data preparation.

\paragraph{Contribution of corruption families.}
We test the simulation of Sec.~\ref{sec:training} by removing
\emph{correlated noise} (flicker, tilt, bias and ray noise),
\emph{wrong-surface observations} (range outliers and observations retained
at an occluder's depth), or \emph{synthetic gaps} (occluders and tracker
dropouts).
Table~\ref{tab:ablation} shows that removing wrong-surface corruptions causes
the largest loss of accuracy across both trackers and geometry models,
supporting their role in learning to repair incorrect observations.
Correlated noise provides smaller mean improvements.
Natural PointOdyssey occlusions remain in every training variant.
Additional synthetic gaps have little effect on overall mean accuracy
but yield small mean gains on occluded points across all four
tracker--geometry combinations (Appendix Table~\ref{tab:ablation_occluded}),
supporting their inclusion during training.

\paragraph{Chunking and track-group size.}
With $C=4$ frames per token, correction is more accurate and about three
times faster than with $C=1$. Increasing to $C=8$ further reduces runtime
but slightly lowers accuracy. We use $P=1024$ tracks per group because
smaller groups reduce accuracy and larger groups offer little improvement
at higher attention cost (Appendix~\ref{app:chunk_group}).

\begin{table}[t]
\centering
\caption{\textbf{Contribution of training corruptions.} Removing wrong-surface corruptions causes the
largest drop. APD3D (\%), averaged over four datasets; gains over the initialized lift are shown
in parentheses, in percentage points. Each
variant is retrained with the same data, network, loss and budget (Appendix~\ref{app:ablation_controls}).}
\label{tab:ablation}
\smallskip
\small
\resizebox{\linewidth}{!}{%
\begin{tabular}{@{}lcccccccc@{}}
\toprule
 & \multicolumn{2}{c}{AllTracker + DA3} & \multicolumn{2}{c}{AllTracker + VGGT-$\Omega$} & \multicolumn{2}{c}{CoWTracker + DA3} & \multicolumn{2}{c}{CoWTracker + VGGT-$\Omega$} \\
\cmidrule(lr){2-3}\cmidrule(lr){4-5}\cmidrule(lr){6-7}\cmidrule(lr){8-9}
Training corruptions & All & Dyn & All & Dyn & All & Dyn & All & Dyn \\
\midrule
Lift (the corrector's input) & 82.1 & 79.3 & 84.6 & 81.6 & 82.2 & 79.5 & 84.9 & 82.3 \\
\midrule
\textbf{Full simulation (ours)} & 83.7 (+1.6) & 81.7 (+2.4) & 86.6 (+2.0) & 84.9 (+3.4) & 84.5 (+2.4) & 82.2 (+2.8) & 87.5 (+2.6) & 85.8 (+3.6) \\
without correlated noise & 83.5 (+1.5) & 81.3 (+2.0) & 86.4 (+1.8) & 84.5 (+2.9) & 84.2 (+2.0) & 82.1 (+2.7) & 87.3 (+2.5) & 85.7 (+3.5) \\
without wrong surface & 82.3 (+0.2) & 79.6 (+0.3) & 84.9 (+0.3) & 82.1 (+0.5) & 82.8 (+0.6) & 80.2 (+0.7) & 85.5 (+0.6) & 83.1 (+0.9) \\
without synthetic gaps & 83.7 (+1.6) & 81.5 (+2.2) & 86.6 (+2.0) & 84.9 (+3.4) & 84.5 (+2.4) & 82.3 (+2.8) & 87.5 (+2.6) & 85.8 (+3.6) \\
white noise only & 82.4 (+0.3) & 79.8 (+0.5) & 85.0 (+0.4) & 82.6 (+1.0) & 82.8 (+0.6) & 80.5 (+1.0) & 85.5 (+0.6) & 83.2 (+1.0) \\
\bottomrule
\end{tabular}}
\end{table}

\FloatBarrier %
\section{Conclusion}
\label{sec:conclusion}

We showed that pretrained 2D tracks lifted with estimated geometry and simple gap initialization
already rival dedicated 3D trackers using the same geometry. Building on this observation,
\method combines these providers with efficient feed-forward trajectory repair and completion
for dense monocular 3D tracking. Training on simulated lifting errors enables one corrector to
improve all evaluated tracker--geometry combinations without retraining, with the largest gains
on occluded points. Our framework achieves state-of-the-art mean accuracy on WorldTrack, while
faster configurations retain competitive accuracy at substantially lower runtime than comparably
accurate 3D trackers. Performance remains sensitive to input geometry (Appendix~\ref{app:kubric}),
motivating future work on robustness to geometric and persistent correspondence errors.

\section*{Acknowledgments}

Yoni Kasten is a Dr. Natalie Shaffer Faculty Fellow and acknowledges support from the Israel Science Foundation (grant 1425/26).
Or Litany acknowledges support from the Israel Science Foundation (grant 624/25) and the Azrieli Foundation Early Career Faculty Fellowship.

\bibliography{refs}
\bibliographystyle{iclr2027_conference}

\clearpage
\appendix

\section{Method Details}
\label{app:method}

\subsection{Observation Preprocessing}
\label{app:observation_preprocessing}

Depth lookup uses the nearest pixel, clamped to the image border, while the viewing ray uses
the sub-pixel track position. The observation mask in Sec.~\ref{sec:lift} uses $\gamma=0.85$
for AllTracker and $\gamma=0.65$ for CoWTracker; the latter is calibrated to match AllTracker's
false-visible rate on 24 PointOdyssey training windows. These thresholds apply to visibility
scores, not tracker confidence outputs.

\subsection{Corrector Implementation}
\label{app:corrector_implementation}

\paragraph{Input and output scaling.}
After scene normalization (Sec.~\ref{sec:representation}), we multiply the initialized displacement,
its masked counterpart and the normalized camera centre in Eq.~\eqref{eq:input} by $S=10$.
The gap feature is $\min(g_{ti},40)/40$. For raw prediction-head output $\mathbf z_{ti}$, the
residual in Eq.~\eqref{eq:output} is $\boldsymbol\Delta_{ti}=\mathbf z_{ti}/S$.
The loss uses scaled errors $S\mathbf e_{ti}$ with unchanged range normalization, giving $S^2\mathcal L$.

\paragraph{Position embeddings.}
For an even embedding width $d$, define
\begin{equation}
  \operatorname{PE}(x;d,\omega)=
  \bigl[\sin(x\,\omega^{-k/(d/2)}),\;\cos(x\,\omega^{-k/(d/2)})\bigr]_{k=0}^{d/2-1}.
  \label{eq:position_encoding}
\end{equation}
The anchor embedding $\boldsymbol\psi(\mathbf a_i)$ concatenates
$\operatorname{PE}(8a_{i,c};32,10^4)$ over the three coordinates $c$, giving 96 features, then applies
Linear$(96\to384)$, GELU and Linear$(384\to384)$. The temporal embedding is
$\mathbf p_n=\operatorname{PE}(n;384,10^3)$, using the chunk index $n$.

\paragraph{Architecture.}
Each block applies temporal self-attention, across-track self-attention and an MLP in sequence,
each with pre-layer normalization and a residual connection.
There is no track-index or pixel embedding, so reordering tracks reorders the outputs.
Predictions can nevertheless depend on which other tracks are included in the group.
Table~\ref{tab:corrector_settings} specifies the architecture. The zero-initialized prediction head
makes the untrained corrector reproduce the initialized trajectories.

\begin{table}[!ht]
\centering
\small
\caption{\textbf{Corrector architecture settings.}}
\label{tab:corrector_settings}
\smallskip
\begin{tabular}{@{}ll@{}}
\toprule
Setting & Value \\
\midrule
Frames per tracklet $C$ & 4 \\
Feature width $D$ & 384 \\
Factorized attention blocks & 8 \\
Attention heads & 6, with width 64 each \\
Feed-forward hidden width / activation & 1536 / GELU \\
Output normalization & Final layer norm \\
Prediction head & Zero-initialized linear map to $3C$ coordinates \\
Parameters & 19.1\,M \\
\bottomrule
\end{tabular}
\end{table}

\subsection{Training Details}
\label{app:training}

\paragraph{Training data and sampling.}
We use 100 scenes (6{,}137 windows) for training and 8 for validation from the PointOdyssey
training split (Sec.~\ref{sec:training}). We exclude \texttt{cab\_e\_ego2}, an alternate view
of WorldTrack's \texttt{cab\_e\_3rd} evaluation recording.
Cached windows contain 84 frames at temporal strides 1 and 2, with starts separated by
42 and 84 source frames, respectively. Eligible tracks have valid trajectories and are visible,
inside the image and in front of the camera in the reference frame; windows with fewer than
1{,}024 are discarded. We cache 2{,}048 tracks per window, at most half moving unless static
tracks are insufficient. A track is marked moving, $\mu_i=1$, if its displacement from its
reference position exceeds 5\,cm anywhere in the cached window; otherwise $\mu_i=0$.

Each training sample draws 512 cached tracks. Each batch retains the first $T$ frames,
with $T$ sampled uniformly from $\{48,68,84\}$, and mirrors trajectory and camera
$x$ coordinates with probability one half. Motion labels remain those of the full cached window.

\paragraph{Corruption processes.}
\label{app:training_corruptions}
The simulated lift in Eq.~\eqref{eq:simulated_lift} uses the log-range perturbation
\begin{equation}
  \varepsilon_{ti}=f_t+\mathbf b_t^{\top}\boldsymbol\delta_{ti}+\beta_{ti}+\eta_{ti}+j_{ti}.
  \label{eq:corrupt}
\end{equation}
Here $f_t$ is frame-shared flicker, $\mathbf b_t\in\mathbb R^2$ a tilt acting on the first two
coordinates $\boldsymbol\delta_{ti}$ of the true ray in reference camera coordinates,
$\beta_{ti}$ a per-track bias, $\eta_{ti}$ white noise and $j_{ti}$ sparse outliers.
The processes $f_t$, $\mathbf b_t$ and $\beta_{ti}$ use unit-variance autoregressive sequences
$z_t=\kappa z_{t-1}+\sqrt{1-\kappa^2}\,\xi_t$, with $z_0,\xi_t\sim\mathcal N(0,1)$,
scaled as in Table~\ref{tab:corruption}. We clip $\varepsilon_{ti}$ to $[-1,1]$ before exponentiation.
To perturb a ray, we project Gaussian noise $\mathcal N(\mathbf 0,\sigma^2 I_3)$ onto its
orthogonal plane, add it to the ray and renormalize.

Synthetic occluders are rectangles moving in the reference camera's normalized image plane
$(x/z,y/z)$; they hide points projected inside and lying behind them.
We remove these observations and contiguous dropouts from ground-truth visibility.
Ground-truth-visible cells hidden by an occluder and adjacent to an available observation can
instead be retained as false-visible. For these cells, we replace the perturbed range by the
clean range times the occluder-to-point depth ratio; angular noise still applies.
We enforce $m_{0i}=1$ and disable occluder-depth replacement in the reference frame.
Table~\ref{tab:corruption} gives the sampling distributions, fixed across training and not
fitted to evaluation sequences.

\begin{table}[h]
\centering
\small
\caption{\textbf{Training corruption distributions.} $\mathcal U$ denotes uniform sampling.
Noise levels and outlier rates are drawn once per training example.}
\label{tab:corruption}
\begin{tabular}{@{}l>{\raggedright\arraybackslash}p{0.68\linewidth}@{}}
\toprule
Term & Distribution \\
\midrule
Flicker $f_t$ & std $\mathcal U(0,0.03)$, autoregressive with $\kappa\sim\mathcal U(0.3,0.95)$ \\
Tilt $\mathbf b_t$ & std $\mathcal U(0,0.03)$, autoregressive with the flicker's $\kappa$ \\
Bias $\beta_{ti}$ & std $\mathcal U(0,0.02)$, autoregressive with $\kappa=0.9$ \\
White noise $\eta_{ti}$ & std $\mathcal U(0.002,0.02)$ \\
Outliers $j_{ti}$ & rate $\mathcal U(0,0.05)$ per cell, next-frame persistence probability $0.5$, jump $\pm\,\mathcal U(0.1,0.5)$ \\
Angular ray noise & std $\mathcal U(0.0005,0.005)$\,rad \\
Occluders & up to two, with independent probabilities $0.6$ and $0.2$; target track drawn in proportion to $\mu_i+10^{-3}$ \\
\quad placement & centred on the target at frame $t_s\sim\mathcal U\{\lfloor T/4\rfloor,\dots,T-1\}$, half-size $\mathcal U(0.03,0.25)$ per axis \\
\quad motion, depth & constant velocity $\mathcal N(0,0.002^2)$ per axis in image-plane units per frame; depth $\mathcal U(0.5,0.9)$ times the target's at $t_s$ \\
Dropouts & $10\%$ of tracks, start $\mathcal U\{1,\dots,T-1\}$, length $\mathcal U\{2,\dots,20\}$ \\
False-visible observations & probability $0.3$ for ground-truth-visible cells hidden by an occluder and adjacent to an available observation \\
\bottomrule
\end{tabular}
\end{table}

\paragraph{Loss settings.}
In Eq.~\eqref{eq:loss}, we set $\rho_{\min}=0.3$, $\lambda=4$ and
$w_{ti}=(1+3\mu_i)(2-m_{ti})$, where $\mu_i$ is the moving-track indicator and $m_{ti}$
the input availability mask. Displacements use consecutive sampled frames, including at temporal stride 2.

\paragraph{Training optimization.}
All runs, including ablations, use 150{,}000 AdamW steps with learning rate $3\cdot10^{-4}$,
weight decay $0.01$ and $\beta=(0.9,0.95)$. A 1{,}000-step linear warm-up is followed by
step-indexed cosine decay to $3\%$ of the peak. Training uses batches of 16, bfloat16 forward
passes and gradient clipping at norm 1 on one RTX PRO~6000 GPU. We evaluate the final
exponential-moving-average checkpoint, with decay $\min(0.999,(1+k)/(10+k))$ at step $k$.

\subsection{Dense Inference}
\label{app:dense_inference}

We correct tracks with positive reference depth; other pixels retain their uncorrected lift.
Eligible tracks are shuffled with seed 0 and processed in groups of $P=1024$, up to 8 groups
per batch (Appendix~\ref{app:chunk_group} evaluates $P$). The shared scale $s$ is the median
reference depth over these eligible pixels; during training, it uses the sampled tracks.

A short final group is filled by cyclically repeating pixels from its batch. Repeated tracks
participate in attention without masking, and their predictions are written to their original
pixel indices. For temporal padding, sequences are extended to a multiple of $C$ with at most
$C-1$ unavailable frames, holding the camera at its last position. Padded-frame predictions
are discarded.

\section{Evaluation Details}
\label{app:evaluation}

\subsection{Evaluation Protocol}
\label{app:protocol}

We evaluate ground-truth points visible and inside the image in the first frame. One ADT clip
has no eligible queries and is excluded for all methods, leaving 199 WorldTrack clips.
Dyn selects complete tracks whose cumulative ground-truth motion over the evaluated frames,
$\sum_{t=0}^{T-2}\|\mathbf x_{t+1,i}-\mathbf x_{ti}\|$, exceeds $0.01$:
metres on WorldTrack and scene units on Kubric.

Following 4RC~\citep{luo20264rc}, RANSAC Sim(3) alignment fits scale, rotation and translation,
whereas global-scale alignment fits only scale. Each alignment is estimated per clip,
separately for All and Dyn. RANSAC uses the same seeds across methods for each clip and scoring set.

Average Jaccard (AJ) and Occlusion Accuracy (OA) follow TAP-Vid~\citep{doersch2022tapvid},
with predicted visibility thresholded at $0.5$.
AJ averages the 3D Jaccard score over the APD3D thresholds of Sec.~\ref{sec:exp_setup},
using the all-point alignment;
OA measures the fraction of point--frames with correct visibility predictions.

Metrics are averaged over clips within each dataset. The WorldTrack mean weights the four
dataset scores equally; Kubric results are averaged over its 50 clips.

\subsection{Baseline Settings}
\label{app:baselines}

Table~\ref{tab:baseline_settings} lists the released-code configurations used for the WorldTrack
comparisons in the main paper and appendix. Kubric uses the same settings with fewer temporal
windows. Native frame sizes are $512^2$ for ADT, $360\times640$ for PS, $720\times1280$ for DR,
and $540\times960$ for PO; the latter three have aspect ratio 16:9.
Predictions are read at query pixels scaled to each method's output grid. Queries outside a crop
use the nearest prediction, except for MotionCrafter on ADT.

\begin{table}[H]
\centering
\caption{\textbf{Baseline configurations.} Input grids are $H\times W$; settings apply to the
accuracy comparisons. Runtime measurements use the settings in Appendix~\ref{app:profiling}.}
\label{tab:baseline_settings}
\smallskip
\begingroup
\small
\setlength{\tabcolsep}{4pt}
\begin{tabular*}{\linewidth}{@{\extracolsep{\fill}}>{\raggedright\arraybackslash}p{0.30\linewidth}cc>{\raggedright\arraybackslash}p{0.36\linewidth}@{}}
\toprule
Method & PS / DR / PO & ADT & Inference settings \\
\midrule
4RC~\citep{luo20264rc} & $280\times504$ & $504^2$ & \\
Any4D~\citep{karhade2025any4d}\newline
V-DPM~\citep{sucar2026vdpm} & $294\times518$ & $518^2$ & \\
St4RTrack~\citep{feng2025st4rtrack}\newline
TraceAnything~\citep{liu2025traceanything} & $288\times512$ & $512^2$ & \\
MotionCrafter~\citep{zhu2026motioncrafter} & $320\times640$ & $320\times640$ & Centre crop. \\
OpenD4RT~\citep{rhos2026opend4rt} & $256^2$ & $256^2$ &
Resize, no crop; stride-2, $128^2$ grid. \\
TrackCraft3R~\citep{nam2026trackcraft3r} & $480\times832$ & $480\times832$ &
Interleaved 12-frame clips, stride 8. \\
Track4World~\citep{lu2026track4world} & $280\times504$ & $512^2$ & \\
DELTAv2~\citep{ngo2025deltav2} & $280\times504$ & $512^2$ & Dense mode. \\
\bottomrule
\end{tabular*}
\endgroup
\end{table}

\paragraph{Evaluation exceptions.}
MotionCrafter is scored on the 63.6\% of ADT queries inside its crop.\footnote{With all ADT queries:
APD3D (All/Dyn) is 60.98/78.51 under Sim(3) and 2.83/5.81 under global scale;
EPE is 0.610\,m under Sim(3); APD2D is 12.33/5.98.}
Affected ADT scores and benchmark means are marked \textsuperscript{\ddag}.
Sim(3) handles the non-metric scale of OpenD4RT and MotionCrafter, and MotionCrafter's
coordinate-frame mismatch with the first camera. Scale-only alignment cannot correct
that mismatch.

\section{Additional Quantitative Results}
\label{app:experiments}

\subsection{Dense Tracking on Kubric}
\label{app:kubric}
Table~\ref{tab:app_kubric} tests dense tracking and sensitivity to geometry providers,
complementing WorldTrack's sparse queries. We use the alignments and WorldTrack
baseline settings in Appendices~\ref{app:protocol} and~\ref{app:baselines}.

Under RANSAC Sim(3), our DA3 and VGGT-$\Omega$ configurations trail 4RC but outperform methods
using the same geometry on all-point APD3D. With 4RC geometry, both tracker variants approach
4RC's all-point accuracy and exceed its dynamic-point accuracy. Under global-scale alignment,
our CoWTracker variant with 4RC geometry leads all three reported metrics.
These results show that transfer across providers remains sensitive to the quality of the input geometry.

\begin{table}[h]
\centering
\caption{\textbf{Dense tracking on Kubric.} TrackCraft3R's~\citep{nam2026trackcraft3r}
50 24-frame Kubric~\citep{greff2022kubric} clips, with ground-truth trajectories for every
reference pixel. APD3D (\%) on all and dynamic tracks and EPE (scene units) under
RANSAC Sim(3) and global-scale alignment; bold best, underline second best.}
\label{tab:app_kubric}
\smallskip
\small
\begin{tabular}{@{}lcccccc@{}}
\toprule
 & \multicolumn{3}{c}{RANSAC Sim(3)} & \multicolumn{3}{c}{Global scale} \\
\cmidrule(lr){2-4}\cmidrule(l){5-7}
Method & All & Dyn & EPE & All & Dyn & EPE \\
\midrule
4RC & \textbf{85.33} & 78.81 & \textbf{0.299} & \underline{82.40} & 73.68 & \underline{0.342} \\
Any4D & 65.31 & 55.51 & 0.854 & 42.69 & 39.77 & 1.107 \\
MotionCrafter & 52.98 & 47.89 & 1.765 & -- & -- & -- \\
OpenD4RT & 38.40 & 37.92 & 1.432 & 24.32 & 26.92 & 1.801 \\
St4RTrack & 59.45 & 54.24 & 3.640 & 56.56 & 51.77 & 3.585 \\
Track4World & 53.55 & 55.53 & 2.074 & 37.37 & 38.82 & 2.552 \\
TraceAnything & 59.29 & 50.40 & 1.023 & 30.46 & 28.02 & 1.252 \\
V-DPM & 75.19 & \textbf{81.03} & 3.369 & 63.70 & 75.18 & 3.405 \\
\midrule
DELTAv2 + DA3 & 63.27 & 62.05 & 1.659 & 44.73 & 43.37 & 2.028 \\
DELTAv2 + VGGT-$\Omega$ & 62.35 & 59.19 & 5.048 & 41.19 & 38.52 & 5.888 \\
TrackCraft3R + DA3 & 62.76 & 55.72 & 1.539 & 41.42 & 38.08 & 2.088 \\
TrackCraft3R + VGGT-$\Omega$ & 59.27 & 50.48 & 4.059 & 34.81 & 31.85 & 6.352 \\
\midrule
\multicolumn{7}{@{}l}{\emph{Ours: the corrector on the lift of}} \\
\textbf{AllTracker + DA3} & 67.10 & 61.32 & 1.794 & 46.72 & 44.01 & 2.144 \\
\textbf{AllTracker + VGGT-$\Omega$} & 65.89 & 59.66 & 4.999 & 43.45 & 39.37 & 5.813 \\
\textbf{AllTracker + 4RC} & 84.28 & 79.79 & 0.336 & 81.99 & \underline{76.23} & 0.347 \\
\textbf{CoWTracker + DA3} & 67.29 & 61.40 & 1.773 & 47.07 & 44.33 & 2.128 \\
\textbf{CoWTracker + VGGT-$\Omega$} & 66.57 & 59.76 & 4.863 & 43.54 & 39.82 & 5.659 \\
\textbf{CoWTracker + 4RC} & \underline{84.88} & \underline{80.66} & \underline{0.324} & \textbf{83.04} & \textbf{77.20} & \textbf{0.335} \\
\bottomrule
\end{tabular}
\end{table}

\subsection{Visibility Metrics and End-Point Error}
\label{app:ajoa}

Tables~\ref{tab:app_ajoa} and~\ref{tab:app_epe} complement APD3D with visibility-aware accuracy
and mean position error. Our CoWTracker variant with 4RC geometry achieves the highest mean AJ
and lowest mean EPE, while CoWTracker with VGGT-$\Omega$ gives the lowest EPE on both real-world
datasets, ADT and PS. Our CoWTracker variants also lead mean OA. Visibility accuracy
comes from the 2D tracker, whose scores we return unchanged (Sec.~\ref{sec:corrector});
the corrector refines the positions used by AJ and EPE.

\begin{table}[t]
\centering
\caption{\textbf{AJ and OA on WorldTrack (\%).} Predicted visibility is thresholded
at $0.5$. AJ uses all-point RANSAC Sim(3) alignment; OA requires no geometric alignment.
Higher is better.}
\label{tab:app_ajoa}
\smallskip
\resizebox{\linewidth}{!}{%
\begin{tabular}{@{}lcccccccccc@{}}
\toprule
 & \multicolumn{2}{c}{ADT} & \multicolumn{2}{c}{PS} & \multicolumn{2}{c}{PO} & \multicolumn{2}{c}{DR} & \multicolumn{2}{c}{Mean} \\
\cmidrule(lr){2-3}\cmidrule(lr){4-5}\cmidrule(lr){6-7}\cmidrule(lr){8-9}\cmidrule(lr){10-11}
Method & AJ & OA & AJ & OA & AJ & OA & AJ & OA & AJ & OA \\
\midrule
OpenD4RT & 58.82 & 91.04 & 69.01 & 87.34 & 45.52 & \textbf{78.63} & 63.04 & 90.03 & 59.10 & 86.76 \\
Track4World & 72.88 & \underline{98.06} & 64.76 & 86.51 & 47.53 & 76.25 & 67.46 & 92.79 & 63.16 & \underline{88.40} \\
\midrule
DELTAv2 + DA3 & 84.18 & 96.94 & 55.18 & 71.57 & 44.90 & 74.35 & 62.74 & 87.36 & 61.75 & 82.56 \\
DELTAv2 + VGGT-$\Omega$ & 83.83 & 95.09 & 57.91 & 71.32 & 45.61 & 74.21 & 68.54 & 87.94 & 63.97 & 82.14 \\
TrackCraft3R + DA3 & 82.95 & 94.26 & 70.18 & 88.13 & 50.29 & 75.59 & 66.79 & 92.69 & 67.55 & 87.67 \\
TrackCraft3R + VGGT-$\Omega$ & 83.48 & 93.48 & 72.40 & \underline{88.29} & 49.72 & 75.64 & 72.68 & 92.92 & 69.57 & 87.58 \\
\midrule
\multicolumn{11}{@{}l}{\emph{Ours: the corrector on the lift of}} \\
\textbf{AllTracker + DA3} & 89.18 & 97.82 & 70.88 & 86.56 & 51.64 & 74.49 & 67.33 & \underline{93.02} & 69.76 & 87.97 \\
\textbf{AllTracker + VGGT-$\Omega$} & 89.75 & 97.82 & 76.17 & 86.56 & 52.21 & 74.49 & 74.27 & \underline{93.02} & 73.10 & 87.97 \\
\textbf{AllTracker + 4RC} & 85.76 & 97.82 & 74.60 & 86.56 & \underline{58.79} & 74.49 & \underline{78.17} & \underline{93.02} & 74.33 & 87.97 \\
\textbf{CoWTracker + DA3} & \underline{89.99} & \textbf{98.51} & 74.34 & \textbf{90.03} & 54.10 & \underline{77.06} & 67.42 & \textbf{93.03} & 71.46 & \textbf{89.66} \\
\textbf{CoWTracker + VGGT-$\Omega$} & \textbf{90.54} & \textbf{98.51} & \textbf{80.13} & \textbf{90.03} & 54.53 & \underline{77.06} & 74.42 & \textbf{93.03} & \underline{74.91} & \textbf{89.66} \\
\textbf{CoWTracker + 4RC} & 86.73 & \textbf{98.51} & \underline{77.93} & \textbf{90.03} & \textbf{61.58} & \underline{77.06} & \textbf{78.34} & \textbf{93.03} & \textbf{76.14} & \textbf{89.66} \\
\bottomrule
\end{tabular}}

\end{table}

\begin{table}[t]
\centering
\caption{\textbf{End-point error on WorldTrack.} Mean 3D position error over evaluated point--frames,
in metres after all-point RANSAC Sim(3) alignment. Lower is better.
\textsuperscript{\ddag}Restricted ADT coverage for MotionCrafter, also affecting its mean
(Appendix~\ref{app:baselines}).}
\label{tab:app_epe}
\smallskip
\small
\begin{tabular}{@{}lccccc@{}}
\toprule
Method & ADT & PS & PO & DR & Mean \\
\midrule
4RC & 0.153 & 0.138 & 0.249 & \underline{0.149} & 0.172 \\
Any4D & 0.286 & 0.216 & 0.335 & 0.191 & 0.257 \\
MotionCrafter & 0.474\textsuperscript{\ddag} & 0.358 & 0.820 & 0.278 & 0.482\textsuperscript{\ddag} \\
OpenD4RT & 0.317 & 0.161 & 0.390 & 0.252 & 0.280 \\
St4RTrack & 0.242 & 0.258 & 0.311 & 0.236 & 0.262 \\
Track4World & 0.269 & 0.184 & 0.366 & 0.219 & 0.260 \\
TraceAnything & 0.198 & 0.284 & 0.735 & 0.378 & 0.399 \\
V-DPM & 0.141 & 0.116 & \textbf{0.179} & 0.189 & \underline{0.156} \\
\midrule
DELTAv2 + DA3 & 0.330 & 0.160 & 0.516 & 0.270 & 0.319 \\
DELTAv2 + VGGT-$\Omega$ & 0.333 & 0.125 & 0.595 & 0.203 & 0.314 \\
TrackCraft3R + DA3 & 0.106 & 0.161 & 0.348 & 0.245 & 0.215 \\
TrackCraft3R + VGGT-$\Omega$ & 0.100 & 0.139 & 0.412 & 0.181 & 0.208 \\
\midrule
\multicolumn{6}{@{}l}{\emph{Ours: the corrector on the lift of}} \\
\textbf{AllTracker + DA3} & 0.089 & 0.151 & 0.344 & 0.242 & 0.207 \\
\textbf{AllTracker + VGGT-$\Omega$} & \underline{0.087} & \underline{0.109} & 0.372 & 0.169 & 0.184 \\
\textbf{AllTracker + 4RC} & 0.114 & 0.119 & 0.266 & \textbf{0.145} & 0.161 \\
\textbf{CoWTracker + DA3} & \underline{0.087} & 0.139 & 0.311 & 0.241 & 0.195 \\
\textbf{CoWTracker + VGGT-$\Omega$} & \textbf{0.085} & \textbf{0.095} & 0.347 & 0.168 & 0.174 \\
\textbf{CoWTracker + 4RC} & 0.110 & 0.110 & \underline{0.231} & \textbf{0.145} & \textbf{0.149} \\
\bottomrule
\end{tabular}

\end{table}

\subsection{Global-Scale Alignment}
\label{app:global}

Table~\ref{tab:main_apd_global} evaluates the predictions of Table~\ref{tab:main_apd} under
global-scale alignment, fitting only scale without adjusting rotation or translation. This tests
whether our conclusions depend on the additional freedom of Sim(3) alignment.
Our configurations using 4RC geometry still lead mean APD3D on both all and dynamic points.
Correction improves all six lifts, adding $1.4$--$2.6$ points on all points and $2.6$--$3.8$ on
dynamic points. Mean accuracy decreases across methods, with substantial drops on PS, where
V-DPM now leads. The mean-accuracy advantage and correction gains therefore persist, while
dataset rankings depend on the alignment used.

\begin{table}[t]
\centering
\caption{\textbf{WorldTrack under global-scale alignment.} APD3D (\%) on all (All) and dynamic
(Dyn) points, fitting only scale (Appendix~\ref{app:protocol}).
MotionCrafter's coordinate-frame mismatch makes its scores non-comparable under this alignment.
\textsuperscript{\ddag}Restricted ADT coverage for MotionCrafter, also affecting its means
(Appendix~\ref{app:baselines}).}
\label{tab:main_apd_global}
\smallskip
\resizebox{\linewidth}{!}{%
\begin{tabular}{@{}lcccccccccc@{}}
\toprule
 & \multicolumn{2}{c}{ADT} & \multicolumn{2}{c}{PS} & \multicolumn{2}{c}{PO} & \multicolumn{2}{c}{DR} & \multicolumn{2}{c}{Mean} \\
\cmidrule(lr){2-3}\cmidrule(lr){4-5}\cmidrule(lr){6-7}\cmidrule(lr){8-9}\cmidrule(lr){10-11}
Method & All & Dyn & All & Dyn & All & Dyn & All & Dyn & All & Dyn \\
\midrule
4RC & 83.09 & 68.24 & 69.15 & 69.15 & 80.23 & 76.44 & 82.95 & 73.13 & 78.86 & 71.74 \\
Any4D & 57.60 & 65.80 & 63.39 & 63.39 & 66.16 & 66.89 & 78.15 & 68.16 & 66.32 & 66.06 \\
MotionCrafter & 2.12\textsuperscript{\ddag} & 6.91\textsuperscript{\ddag} & 0.00 & 0.00 & 0.11 & 0.02 & 1.59 & 1.39 & 0.96\textsuperscript{\ddag} & 2.08\textsuperscript{\ddag} \\
OpenD4RT & 69.28 & 67.96 & \underline{76.87} & \underline{76.87} & 65.19 & 70.39 & 74.80 & \underline{76.25} & 71.53 & 72.87 \\
St4RTrack & 74.33 & 71.13 & 69.12 & 69.12 & 67.86 & 68.34 & 74.10 & 67.31 & 71.35 & 68.97 \\
Track4World & 75.63 & 85.43 & 71.39 & 71.39 & 63.93 & 65.61 & 70.74 & 68.12 & 70.42 & 72.64 \\
TraceAnything & 73.64 & 67.34 & 67.60 & 67.60 & 38.44 & 44.02 & 60.85 & 58.59 & 60.13 & 59.39 \\
V-DPM & 84.41 & 69.61 & \textbf{78.40} & \textbf{78.40} & \underline{81.43} & \textbf{82.19} & 75.56 & \textbf{78.13} & 79.95 & 77.08 \\
\midrule
DELTAv2 + DA3 & 78.63 & 85.85 & 65.05 & 65.05 & 54.83 & 56.57 & 69.41 & 57.51 & 66.98 & 66.25 \\
DELTAv2 + VGGT-$\Omega$ & 79.43 & 85.12 & 70.50 & 70.50 & 58.61 & 60.48 & 76.19 & 68.11 & 71.18 & 71.05 \\
TrackCraft3R + DA3 & 89.12 & 84.04 & 63.91 & 63.91 & 65.51 & 62.25 & 71.26 & 60.14 & 72.45 & 67.59 \\
TrackCraft3R + VGGT-$\Omega$ & 90.46 & 83.55 & 68.13 & 68.13 & 67.72 & 64.21 & 77.69 & 70.43 & 76.00 & 71.58 \\
\midrule
\multicolumn{11}{@{}l}{\emph{Ours: the corrector on the lift of}} \\
\textbf{AllTracker + DA3} & 90.71 & \underline{91.40} & 64.80 & 64.80 & 70.22 & 70.71 & 72.49 & 61.92 & 74.55 & 72.21 \\
\textbf{AllTracker + VGGT-$\Omega$} & \underline{91.93} & 88.14 & 71.38 & 71.38 & 72.65 & 73.10 & 81.15 & 74.68 & 79.28 & 76.82 \\
\textbf{AllTracker + 4RC} & 89.22 & 90.71 & 71.31 & 71.31 & 80.42 & 76.78 & \textbf{84.63} & 75.18 & \underline{81.40} & \underline{78.49} \\
\textbf{CoWTracker + DA3} & 90.73 & \textbf{91.90} & 65.93 & 65.93 & 72.12 & 72.93 & 72.33 & 62.17 & 75.28 & 73.23 \\
\textbf{CoWTracker + VGGT-$\Omega$} & \textbf{92.32} & 88.82 & 72.49 & 72.49 & 74.15 & 74.62 & 80.82 & 74.10 & 79.94 & 77.51 \\
\textbf{CoWTracker + 4RC} & 89.08 & 91.30 & 72.96 & 72.96 & \textbf{82.27} & \underline{78.82} & \underline{84.12} & 75.80 & \textbf{82.11} & \textbf{79.72} \\
\bottomrule
\end{tabular}}

\end{table}

\subsection{Image-Plane Accuracy}
\label{app:apd2d}

Table~\ref{tab:main_apd2d} complements the 3D evaluation with the Average Percentage of 2D Points
within Delta (APD2D), averaging accuracy over $\{1,2,4,8,16\}$-pixel thresholds after rescaling
coordinates to $256\times256$. We use the 84-frame clips, initial queries, All/Dyn definitions
and aggregation of Appendix~\ref{app:protocol}.

\paragraph{Evaluation protocol.}
We project 3D predictions through their associated cameras without ground-truth alignment.
Projections are mapped to original image coordinates, accounting for resizing and crop offsets,
before the common rescaling. Point--frames behind the ground-truth camera are excluded;
predictions behind their camera or with non-finite projections count as misses at every threshold.
Native AllTracker and CoWTracker predictions are scored directly in 2D at the same queries,
with the same scaling and ground-truth masks.
Track4World and DELTAv2 use inputs capped to our grid, with ADT uncapped
(Appendix~\ref{app:baselines}). PS has identical All and Dyn scores because all its queries
meet the dynamic-track criterion.

\paragraph{Cameras for APD2D.}
Table~\ref{tab:apd2d_cameras} lists the projection cameras. Our method, TrackCraft3R and DELTAv2 use their geometry
provider's predicted intrinsics and extrinsics, following the shared estimated-geometry setup in
Sec.~\ref{sec:exp_setup}. For methods without camera outputs, cameras are recovered from their
own predictions without ground-truth camera parameters.

\begin{table}[t]
\centering
\caption{\textbf{Camera sources for APD2D.} Cameras are predicted, supplied by a geometry provider,
or recovered from model outputs; no ground-truth camera parameters are used.}
\label{tab:apd2d_cameras}
\smallskip
\small
\begin{tabular}{@{}p{0.43\linewidth}p{0.52\linewidth}@{}}
\toprule
Methods & Projection cameras \\
\midrule
4RC, Any4D, Track4World, V-DPM & Predicted by the method \\
TrackCraft3R, DELTAv2 & Input geometry cameras \\
Our configurations & Predicted by the geometry provider \\
MotionCrafter & Robust joint pose/focal fit to predicted point maps \\
St4RTrack, OpenD4RT, TraceAnything & Recovered from each method's own outputs \\
\bottomrule
\end{tabular}
\end{table}

\begin{table}[t]
\centering
\caption{\textbf{Image-plane accuracy on WorldTrack.} APD2D (\%) on all (All) and dynamic (Dyn) points.
3D trajectories are projected through the cameras in Table~\ref{tab:apd2d_cameras};
native 2D references are scored without projection. Bold and underline rank 3D methods only.
Scoring details are in Appendix~\ref{app:apd2d}.
\textsuperscript{\ddag}Restricted ADT coverage for MotionCrafter, also affecting its means
(Appendix~\ref{app:baselines}).}
\label{tab:main_apd2d}
\smallskip
\resizebox{\linewidth}{!}{%
\begin{tabular}{@{}lcccccccccc@{}}
\toprule
 & \multicolumn{2}{c}{ADT} & \multicolumn{2}{c}{PS} & \multicolumn{2}{c}{PO} & \multicolumn{2}{c}{DR} & \multicolumn{2}{c}{Mean} \\
\cmidrule(lr){2-3}\cmidrule(lr){4-5}\cmidrule(lr){6-7}\cmidrule(lr){8-9}\cmidrule(lr){10-11}
Method & All & Dyn & All & Dyn & All & Dyn & All & Dyn & All & Dyn \\
\midrule
4RC & 62.06 & 27.36 & 51.79 & 51.79 & 71.50 & 60.91 & 79.15 & 65.26 & 66.13 & 51.33 \\
Any4D & 40.90 & 23.60 & 31.46 & 31.46 & 56.99 & 44.39 & 76.93 & 50.10 & 51.57 & 37.39 \\
MotionCrafter & 17.47\textsuperscript{\ddag} & 10.34\textsuperscript{\ddag} & 15.11 & 15.11 & 21.88 & 13.70 & 40.69 & 30.58 & 23.79\textsuperscript{\ddag} & 17.43\textsuperscript{\ddag} \\
OpenD4RT & 45.94 & 21.77 & 46.90 & 46.90 & 56.13 & 49.43 & 72.58 & 57.41 & 55.39 & 43.88 \\
St4RTrack & 59.53 & 28.47 & 39.12 & 39.12 & 63.95 & 47.47 & 79.77 & 50.73 & 60.59 & 41.45 \\
Track4World & 76.31 & \textbf{66.42} & 63.67 & 63.67 & \underline{76.82} & \textbf{68.95} & \textbf{89.73} & \textbf{74.30} & 76.63 & \textbf{68.34} \\
TraceAnything & 23.13 & 11.51 & 32.30 & 32.30 & 35.29 & 28.14 & 59.93 & 41.65 & 37.66 & 28.40 \\
V-DPM & 50.15 & 22.88 & 55.04 & 55.04 & \textbf{77.00} & 64.66 & 85.10 & 64.42 & 66.82 & 51.75 \\
\midrule
DELTAv2 + DA3 & 76.27 & \underline{65.59} & 66.20 & 66.20 & 72.94 & 61.94 & 87.28 & 66.65 & 75.68 & 65.09 \\
DELTAv2 + VGGT-$\Omega$ & 74.75 & 62.12 & 66.15 & 66.15 & 72.52 & 62.32 & 87.52 & 67.33 & 75.23 & 64.48 \\
TrackCraft3R + DA3 & 48.89 & 20.04 & 42.44 & 42.44 & 63.29 & 45.66 & 75.85 & 54.15 & 57.62 & 40.57 \\
TrackCraft3R + VGGT-$\Omega$ & 52.14 & 22.21 & 41.59 & 41.59 & 61.10 & 44.52 & 77.27 & 55.37 & 58.02 & 40.92 \\
\midrule
\multicolumn{11}{@{}l}{\emph{Ours (2D tracker + geometry provider)}} \\
\textbf{AllTracker + DA3} & 76.59 & 60.36 & 63.41 & 63.41 & 74.51 & 66.05 & 86.35 & 70.14 & 75.22 & 64.99 \\
\textbf{AllTracker + VGGT-$\Omega$} & 76.47 & 59.19 & 63.56 & 63.56 & 74.33 & 65.92 & 86.67 & 70.88 & 75.26 & 64.89 \\
\textbf{AllTracker + 4RC} & 76.45 & 60.84 & 63.66 & 63.66 & 74.68 & 66.29 & 86.69 & 71.00 & 75.37 & 65.45 \\
\textbf{CoWTracker + DA3} & \textbf{76.98} & 62.64 & 67.18 & 67.18 & 76.31 & 67.41 & 88.08 & 72.15 & \underline{77.14} & 67.35 \\
\textbf{CoWTracker + VGGT-$\Omega$} & 76.78 & 61.43 & \underline{67.23} & \underline{67.23} & 76.11 & 67.28 & 88.35 & 72.88 & 77.12 & 67.21 \\
\textbf{CoWTracker + 4RC} & \underline{76.91} & 63.23 & \textbf{67.35} & \textbf{67.35} & 76.41 & \underline{67.62} & \underline{88.39} & \underline{72.93} & \textbf{77.27} & \underline{67.78} \\
\midrule
\multicolumn{11}{@{}l}{\emph{Native 2D references (no 3D prediction)}} \\
AllTracker (2D tracker alone) & 77.18 & 67.90 & 64.29 & 64.29 & 75.00 & 68.68 & 88.31 & 73.55 & 76.19 & 68.60 \\
CoWTracker (2D tracker alone) & 78.32 & 71.22 & 68.59 & 68.59 & 77.44 & 70.58 & 90.67 & 76.00 & 78.75 & 71.60 \\
\bottomrule
\end{tabular}}

\end{table}

\paragraph{Results.}
Native CoWTracker achieves the highest APD2D in every reported column, outperforming projected
3D trajectories and motivating our use of strong pretrained 2D correspondences
(Sec.~\ref{sec:exp_apd2d}).
Our CoWTracker variant with 4RC geometry leads mean All APD2D among 3D methods and ranks second
on Dyn, behind Track4World. Switching from AllTracker to CoWTracker improves every dataset and
geometry pairing, adding 1.86--1.92 points to mean All and 2.32--2.36 to mean Dyn. For either
tracker, the means vary across geometry providers by only 0.15 (All) and at most 0.57 (Dyn) points.
Tracker choice therefore has a larger effect on APD2D here.

\paragraph{Relation to 3D accuracy.}
All six of our configurations remain below their native tracker in both means.
With CoWTracker and 4RC geometry, the gaps are 1.48 points
on All and 3.82 on Dyn. This comparison evaluates the complete pipeline; it neither isolates
the corrector's effect nor separates visible and occluded points.

APD2D evaluates trajectories together with their projection cameras, including any camera recovery.
It cannot measure range errors along a valid viewing ray: lifting and reprojecting through the
same camera returns the input pixel. The small variation across geometry providers therefore
does not imply similar 3D accuracy. The APD3D gains over initialized lifts in Fig.~\ref{fig:gain}
provide complementary evidence of improved 3D estimation.

\section{Ablations and Efficiency}
\label{app:ablations}
\suppressfloats[t] %

\subsection{Training Ablation Settings}
\label{app:ablation_controls}

\paragraph{Simulated versus provider-generated lifts.}
Provider-generated lifts use AllTracker with DA3 or VGGT-$\Omega$.
To account for differences in coordinate frame and scale, we align
ground-truth frame-0 positions $\mathbf x_{0i}$ to lifted positions
$\mathbf y_{0i}$ using a similarity transform
$\alpha R\mathbf x_{0i}+\mathbf b$.
Valid points have positive frame-0 range.
We first fit the transform by least squares on all valid points, then
perform three refits to reduce the influence of outliers.
After each fit, we compute the relative residual
\[
  e_i =
  \frac{\|\alpha R\mathbf x_{0i}+\mathbf b-\mathbf y_{0i}\|}
       {\|\mathbf y_{0i}\|}.
\]
The next fitting set comprises valid points satisfying
$e_i\leq\max(3\,\mathrm{median}_{j\in K}e_j,\,0.01)$,
where $K$ is the set used for the current fit.
The $1\%$ floor limits trimming when residuals are already low.
After the final fit, we recompute the median residual over all valid points,
including those excluded from fitting.
We retain windows with at least 256 valid points, positive fitted scale,
and a finite median residual below $0.2$.

We apply this single transform to the ground truth at every frame,
leaving the input lift unchanged.
There is no per-frame refitting or additional alignment inside the loss,
so errors in the estimated camera motion remain part of the correction task.
After filtering, the runs trained on AllTracker+DA3 and
AllTracker+VGGT-$\Omega$ lifts use 3{,}890 and 3{,}837
training windows, respectively. The latter must pass both providers' filters.

These models improve training-scene APD3D on all points by 14.14 and 8.33
percentage points over their corresponding lifts, respectively.

\begin{figure}[!htbp]
\centering
\includegraphics[width=0.95\linewidth]{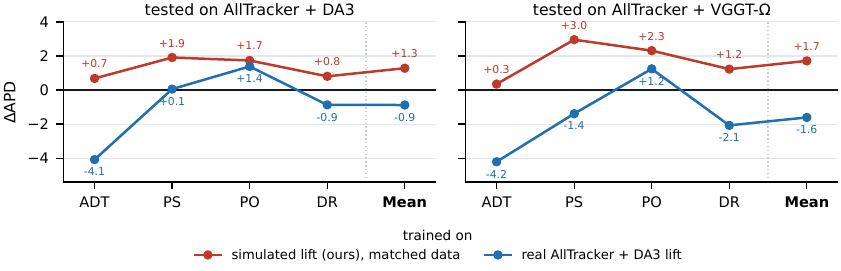}
\caption{\textbf{Simulated versus AllTracker+DA3 training lifts with matched data.}
Both models train on the same 3{,}890 filtered windows for 150{,}000 steps.
APD3D gains on all points are reported in percentage points over the
initialized lift, with AllTracker and DA3 (left) or VGGT-$\Omega$ (right).
``Mean'' averages the four datasets.}
\label{fig:regime_matched_data}
\end{figure}

We train an additional simulated-lift model on the same 3{,}890 filtered
windows as the model trained on AllTracker+DA3 lifts, with the same
architecture, optimization settings and 150{,}000 training steps.
Figure~\ref{fig:regime_matched_data} shows that simulated training still
improves APD3D on every dataset and outperforms training on AllTracker+DA3
lifts with both test geometry providers.

\paragraph{Corruption families.}
The variants in Table~\ref{tab:ablation} are retrained with the same data,
network, loss and optimization settings, removing one corruption family
or retaining only white noise among the added perturbations.
Natural PointOdyssey occlusions remain in all variants, so completion
supervision is present even without synthetic gaps.

Table~\ref{tab:ablation_occluded} compares these variants on
ground-truth-occluded point--frames.
Full simulation achieves slightly higher mean APD3D than the variant
without synthetic gaps across all four tracker--geometry combinations.
This comparison supports a modest benefit from adding synthetic gaps
to the occlusions already present in the training data.
Removing synthetic occluders also removes their false-visible observations
at the occluder's depth, so this ablation does not isolate the effect
of missing observations alone.

\begin{table}[t]
\centering
\caption{\textbf{Contribution of training corruptions on occluded points.}
APD3D (\%) on ground-truth-occluded point--frames, averaged over
the four datasets.
Scoring retains the all-point RANSAC Sim(3) alignment.
Parentheses report percentage-point gains over the initialized lift.}
\label{tab:ablation_occluded}
\smallskip
\small
\begin{tabular}{@{}lcccc@{}}
\toprule
 & \multicolumn{2}{c}{AllTracker}
 & \multicolumn{2}{c}{CoWTracker} \\
\cmidrule(lr){2-3}\cmidrule(lr){4-5}
Training corruptions
 & DA3 & VGGT-$\Omega$ & DA3 & VGGT-$\Omega$ \\
\midrule
Lift (the corrector's input)
 & 71.9 & 73.7 & 70.0 & 71.6 \\
\midrule
\textbf{Full simulation (ours)}
 & \textbf{75.7 (+3.8)} & \textbf{79.0 (+5.2)}
 & \textbf{77.1 (+7.1)} & \textbf{80.8 (+9.2)} \\
without correlated noise
 & 75.3 (+3.4) & 78.7 (+5.0)
 & 76.7 (+6.7) & 80.2 (+8.6) \\
without wrong surface
 & 72.7 (+0.8) & 75.3 (+1.5)
 & 71.9 (+1.9) & 74.0 (+2.4) \\
without synthetic gaps
 & 75.6 (+3.7) & 78.9 (+5.1)
 & 77.0 (+7.0) & 80.6 (+8.9) \\
white noise only
 & 72.3 (+0.4) & 74.8 (+1.1)
 & 71.4 (+1.4) & 73.3 (+1.7) \\
\bottomrule
\end{tabular}
\end{table}

\subsection{Ray Masking}
\label{app:ray_masking}

Table~\ref{tab:app_ray_masking} evaluates the ray-masking choice in Sec.~\ref{sec:lift}.
Masking improves ADT APD3D by about $2.8$ points overall and $8.1$ on occluded points
with either geometry provider. Changes elsewhere are below one point and mixed.

\begin{table}[!hbp]
\centering
\caption{\textbf{Ray masking.} APD3D (\%) for separately trained correctors that mask or retain
rays where the input mask is zero, using AllTracker and $C=4$. All and Occ share the all-point
RANSAC Sim(3) alignment; Occ scores ground-truth-occluded point--frames.}
\label{tab:app_ray_masking}
\smallskip
\begingroup
\small
\setlength{\tabcolsep}{4pt}
\begin{tabular}{@{}ll*{8}{c}@{}}
\toprule
 & & \multicolumn{2}{c}{ADT} & \multicolumn{2}{c}{PS} & \multicolumn{2}{c}{PO} & \multicolumn{2}{c}{DR} \\
\cmidrule(lr){3-4}\cmidrule(lr){5-6}\cmidrule(lr){7-8}\cmidrule(l){9-10}
Geometry & Subset & \shortstack{With\\masking} & \shortstack{Without\\masking} & \shortstack{With\\masking} & \shortstack{Without\\masking} & \shortstack{With\\masking} & \shortstack{Without\\masking} & \shortstack{With\\masking} & \shortstack{Without\\masking} \\
\midrule
DA3 & All & 93.69 & 90.93 & 85.46 & 85.44 & 76.04 & 76.03 & 79.50 & 79.52 \\
DA3 & Occ & 90.27 & 82.22 & 76.08 & 76.30 & 73.19 & 73.03 & 63.17 & 64.05 \\
VGGT-$\Omega$ & All & 94.04 & 91.23 & 90.40 & 90.36 & 76.10 & 76.10 & 85.74 & 85.75 \\
VGGT-$\Omega$ & Occ & 90.12 & 82.03 & 82.04 & 82.54 & 72.49 & 72.23 & 71.26 & 71.85 \\
\bottomrule
\end{tabular}
\endgroup
\end{table}

\subsection{Chunk Size and Group Size}
\label{app:chunk_group}

Table~\ref{tab:app_chunk} compares separately trained correctors with $C\in\{1,4,8\}$,
each trained for 150{,}000 steps. Our setting, $C=4$, gives the highest accuracy;
$C=8$ reduces correction time by about a third at a cost of at most $0.3$ APD3D points.
Without chunking ($C=1$), correction is slower and less accurate under this training setup.

\begin{table}[t]
\centering
\caption{\textbf{Temporal chunking and pipeline runtime.} AllTracker with DA3 or VGGT-$\Omega$
geometry. \textbf{Left:} correctors trained separately for each chunk size $C$ (frames per token),
reporting mean WorldTrack APD3D (\%) under RANSAC Sim(3) alignment and corrector-only runtime
at $512^2$. \textbf{Right:} pipeline runtime by stage across resolutions, using our default $C=4$.
All timings use 84 frames on one RTX PRO~6000; profiling details appear in Appendix~\ref{app:profiling}.}
\label{tab:app_chunk}
\smallskip
\begin{minipage}[c]{0.58\linewidth}
\centering
\small
\begin{tabular}{@{}lcccccc@{}}
\toprule
 & \multicolumn{2}{c}{DA3 input} & \multicolumn{2}{c}{VGGT-$\Omega$ input} & \multicolumn{2}{c}{Corrector time (s)} \\
\cmidrule(lr){2-3}\cmidrule(lr){4-5}\cmidrule(l){6-7}
$C$ & All & Dyn & All & Dyn & DA3 & VGGT-$\Omega$ \\
\midrule
1 & 82.93 & 81.08 & 85.83 & 83.79 & 10.0 & 9.7 \\
\textbf{4} & 83.68 & 81.74 & 86.57 & 84.92 & 3.1 & 3.0 \\
8 & 83.59 & 81.48 & 86.48 & 84.75 & 2.0 & 1.9 \\
\bottomrule
\end{tabular}
\end{minipage}\hfill
\begin{minipage}[c]{0.40\linewidth}
\centering
\includegraphics[width=\linewidth]{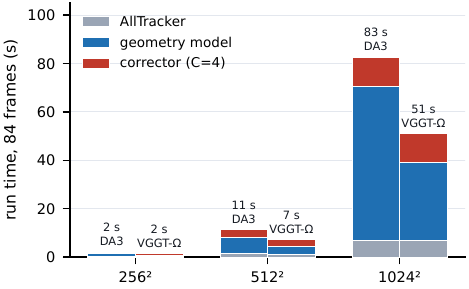}
\end{minipage}
\end{table}

\begin{figure}[t]
\centering
\includegraphics[trim=190bp 0bp 0bp 10bp,clip,width=0.65\linewidth]{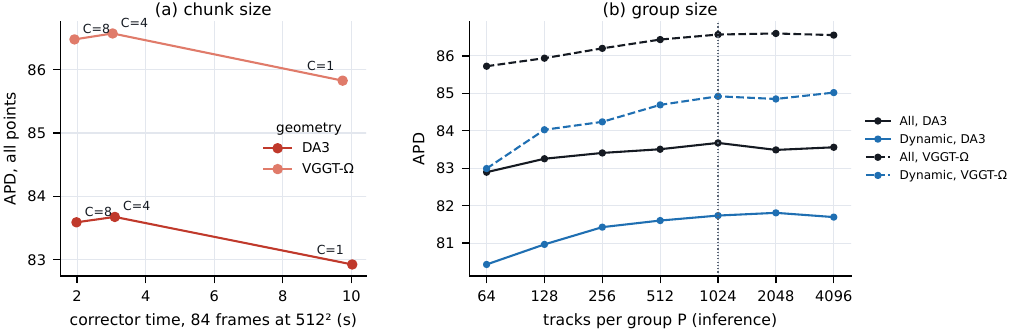}
\caption{\textbf{Group size at inference.} Mean WorldTrack APD3D (\%) on all and dynamic points
under RANSAC Sim(3) alignment, varying tracks per group $P$ with AllTracker and fixed corrector
weights ($C=4$). The dotted line marks our setting, $P=1024$.}
\label{fig:group_size}
\end{figure}

Unlike chunk size, group size varies only at inference, with the checkpoint fixed
(Fig.~\ref{fig:group_size}). Smaller groups reduce accuracy, particularly on dynamic points;
increasing $P$ from 1024 to 4096 changes every score by at most $0.2$ points.
We therefore use $P=1024$.
For $M$ total tracks and $N$ temporal chunks, the across-track attention cost in
Sec.~\ref{sec:corrector} sums to $O(MNP)$ over all groups, increasing linearly with $P$.
Complete pipeline timings are reported in Appendix~\ref{app:profiling}.

\subsection{Run Time and Memory}
\label{app:profiling}

Table~\ref{tab:app_profiling} reports complete pipeline costs for 84 frames from one Dynamic Replica
clip at three nominal resolutions on a 96\,GB RTX PRO~6000 GPU, using each method's released
inference code. Timing includes all models (the 2D tracker, geometry model and corrector for ours)
and runs from decoded frames in host memory to dense trajectories returned to host memory;
video decoding and model loading are excluded. At $256^2$ and $512^2$, we report the median of three runs after
one warm-up; $1024^2$ uses a single run without warm-up. For separately profiled stages, we sum
their median times. Memory is the maximum GPU allocation across stages and timed runs.
Input sizes follow each method's patch-size rounding, with resolution overrides marked in the table.

Our VGGT-$\Omega$ configurations offer competitive WorldTrack accuracy at substantially lower
runtime than 4RC and V-DPM, while faster baselines are less accurate (Fig.~\ref{fig:teaser}).
The stage breakdown in Table~\ref{tab:app_chunk} (right) connects these measurements to grouped
inference (Sec.~\ref{sec:inference}). With AllTracker, correction takes roughly 3\,s at $512^2$
and 12\,s at $1024^2$, consistent with linear scaling in pixel count. Geometry estimation grows
faster, so the complete pipeline does not share this scaling. Our AllTracker configurations with
DA3 or VGGT-$\Omega$ remain feasible at $1024^2$, where 4RC, V-DPM and DELTAv2 run out of memory.

\begin{table}[h]
\centering
\caption{\textbf{Run time and peak allocated GPU memory.} Complete pipelines at three nominal
resolutions. OOM: out of memory; FAIL: St4RTrack's
upsampling exceeds an implementation tensor-size limit; --: not run.
$^\dagger$Hard-coded input resolution overridden for this experiment.
OpenD4RT runs only at $256^2$; ours uses $C=4$.}
\label{tab:app_profiling}
\smallskip
\resizebox{\linewidth}{!}{%
\begin{tabular}{@{}lcccccc@{}}
\toprule
 & \multicolumn{3}{c}{Run time (s)} & \multicolumn{3}{c}{Peak memory (GiB)} \\
\cmidrule(lr){2-4}\cmidrule(l){5-7}
Method & $256^2$ & $512^2$ & $1024^2$ & $256^2$ & $512^2$ & $1024^2$ \\
\midrule
\multicolumn{7}{@{}l}{\emph{Ours: the corrector on the lift of}} \\
\textbf{AllTracker + DA3} & 2.1 & 11.4 & 82.6 & 10.1 & 21.3 & 63.7 \\
\textbf{AllTracker + VGGT-$\Omega$} & 1.5 & 7.3 & 51.0 & 7.9 & 12.5 & 43.9 \\
\textbf{CoWTracker + DA3} & 2.8 & 15.3 & OOM & 10.6 & 36.5 & OOM \\
\textbf{CoWTracker + VGGT-$\Omega$} & 2.2 & 11.7 & OOM & 10.6 & 36.5 & OOM \\
\midrule
4RC & 4.3 & 31.8 & OOM & 11.0 & 26.8 & OOM \\
V-DPM$^\dagger$ & 77.3 & 590.8 & OOM & 26.9 & 85.7 & OOM \\
TrackCraft3R + DA3 / VGGT-$\Omega$ & 5.0 / 4.4 & 22.1 / 18.5 & 162.7 / 132.3 & 9.8 / 7.6 & 21.0 / 12.2 & 63.4 / 30.9 \\
DELTAv2 + DA3 / VGGT-$\Omega$ & 7.8 / 7.2 & 33.4 / 29.8 & OOM / OOM & 9.8 / 8.3 & 32.5 / 32.5 & OOM / OOM \\
Any4D$^\dagger$ & 0.6 & 3.2 & OOM & 13.7 & 45.2 & OOM \\
TraceAnything$^\dagger$ & 1.9 & 5.2 & 20.4 & 3.0 & 4.8 & 11.8 \\
Track4World & 1.5 & 8.4 & 85.7 & 10.1 & 21.1 & 70.2 \\
St4RTrack & 0.6 & 2.9 & FAIL & 6.8 & 20.8 & FAIL \\
MotionCrafter$^\dagger$ & 4.8 & 20.4 & OOM & 10.9 & 29.1 & OOM \\
OpenD4RT ($256^2$ only) & 284.2 & -- & -- & 7.8 & -- & -- \\
\bottomrule
\end{tabular}}
\end{table}

\FloatBarrier
\section{Additional Qualitative Results}
\label{app:qualitative}

\begin{figure}[!hbp]
\centering
\includegraphics[width=\linewidth]{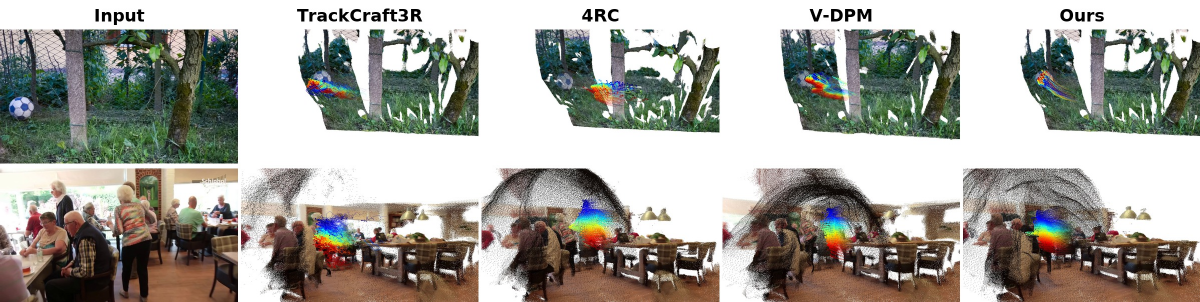}
\caption{\textbf{Additional qualitative comparisons.} Top: DAVIS \emph{soccerball}; bottom:
\emph{lady-running}. Ours uses CoWTracker with VGGT-$\Omega$ (top) and AllTracker with DA3 (bottom),
with the same corrector weights ($C=4$). TrackCraft3R uses VGGT-$\Omega$ in both rows
(unlike ours in \emph{lady-running}); 4RC and V-DPM use their own reconstructions.
Methods track the same reference pixels, shown on their respective reconstructions from matched
novel viewpoints.}
\label{fig:qualitative_appendix_cmp}
\end{figure}

\paragraph{Additional method comparisons.}
Figure~\ref{fig:qualitative_appendix_cmp} adds two DAVIS~\citep{perazzi2016benchmark} examples.
In \emph{soccerball}, which includes occlusion by a tree trunk, our trajectories form cleaner arcs
ending on the ball. In \emph{lady-running}, ours follow the body in an ordered band, while
TrackCraft3R's scatter below it; 4RC and V-DPM also produce compact trajectories.

\begin{figure}[!htbp]
\centering
\includegraphics[width=\linewidth]{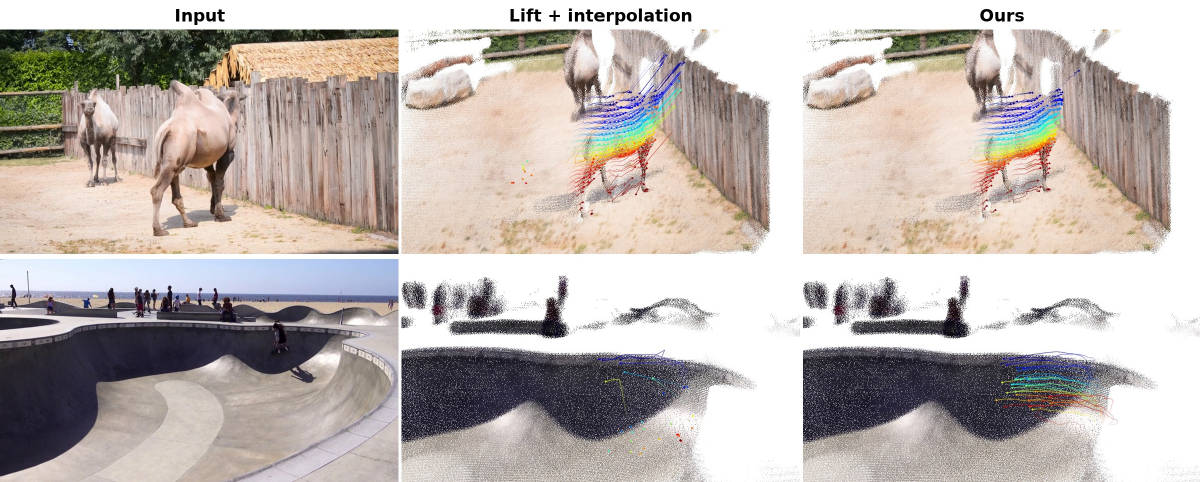}
\caption{\textbf{Effect of trajectory correction.} Left to right: input frame, initialized lift,
and corrected trajectories ($C=4$). Top: DAVIS \emph{camel}; bottom: \emph{skate-park}.
CoWTracker's 2D tracks and DA3 geometry are held fixed; both trajectory panels use the same
point cloud and novel viewpoint.}
\label{fig:qualitative_appendix}
\end{figure}

\paragraph{Isolating trajectory repair.}
Figure~\ref{fig:qualitative_appendix} isolates the effect of correction on the initialized lift
(Sec.~\ref{sec:representation}). In \emph{camel}, correction reduces trajectory outliers near the
ground and produces more coherent paths along the body. In \emph{skate-park}, scattered trajectories
become a more coherent bundle. These comparisons complement the quantitative gains over the
initialized lift in Fig.~\ref{fig:gain}.

\FloatBarrier
\paragraph{Visualization details.}
Figures~\ref{fig:qualitative}, \ref{fig:qualitative_appendix_cmp}
and~\ref{fig:qualitative_appendix} use queries spaced 16 pixels apart at native resolution
inside the eroded first-frame object mask. Colors identify queries in raster order; trails show
the current and preceding 20 frames, with dots marking current positions. Backdrops combine
confident background points from all frames, outside the dilated object mask, with object points
from the displayed frame. The method comparisons show each clip's third-to-last frame.
4RC and V-DPM outputs are read at the nearest pixels of their grids; TrackCraft3R uses its
interleaved benchmark mode with queries lifted using VGGT-$\Omega$ depth.
To match visualization viewpoints across methods, a novel camera in VGGT-$\Omega$ coordinates
is mapped into each reconstruction by a similarity fitted to first-frame point maps,
trimming 20\% outliers over two rounds.

\end{document}